\documentclass{article} 
\usepackage{iclr2027_conference,times}

\usepackage{amsmath,amsfonts,bm}

\def\eqref#1{equation~\ref{#1}}

\def\1{\bm{1}}

\DeclareMathAlphabet{\mathsfit}{\encodingdefault}{\sfdefault}{m}{sl}
\SetMathAlphabet{\mathsfit}{bold}{\encodingdefault}{\sfdefault}{bx}{n}

\usepackage{hyperref}
\usepackage{url}
\usepackage{microtype}
\usepackage{pgfplots}
\usepgfplotslibrary{groupplots}
\usepackage{amsthm}
\usepackage{cleveref}
\usepackage{wrapfig}

\makeatletter
\newcommand{\newreptheorem}[2]{%
\newenvironment{rep#1}[1]{%
 \def\rep@title{#2 \ref{##1}}%
 \begin{rep@theorem}}%
 {\end{rep@theorem}}}
\makeatother

\newreptheorem{theorem}{Theorem}
\newreptheorem{lemma}{Lemma}

\newcommand{\rone}{(\emph{i})~}
\newcommand{\rtwo}{(\emph{ii})~}
\newcommand{\rthree}{(\emph{iii})~}

\newcommand{\lbbar}{\{\kern-0.5ex|}
\newcommand{\rbbar}{|\kern-0.5ex\}}

\newcommand{\biglbbar}{\left\{\kern-0.5ex\left|}
\newcommand{\bigrbbar}{\right|\kern-0.5ex\right\}}

\newcommand{\Omit}[1]{}

\newcounter{resq}

\definecolor{dgreen}{RGB}{0,128,0}
\definecolor{dred}{RGB}{200,0,0}

\theoremstyle{definition}
\newtheorem{example}{Example}
\theoremstyle{plain}

\newcommand{\sample}{x}
\newcommand{\reward}{R}
\newcommand{\param}{\theta}
\newcommand{\rreward}{R'_{\param}}
\newcommand{\policy}{p_{\param}}
\newcommand{\bpolicy}{\pi_{\param}}
\newcommand{\qpolicy}{q_{\param}}
\newcommand{\p}{\mathbb{P}}
\newcommand{\e}{\mathbb{E}}

\newcommand{\cweight}{W}
\newcommand{\bweight}{B}

\newcommand{\norm}[1]{\left\lVert #1 \right\rVert}
\newcommand{\kld}{D_{\mathrm{KL}}}

\definecolor{plotcorrected}{HTML}{008000}
\definecolor{plotuncorrected}{HTML}{E08214}
\definecolor{plotunenriched}{HTML}{0072B2}
\definecolor{plotgaussian}{HTML}{009E73}
\definecolor{plotscalelow}{HTML}{8C568A}
\definecolor{plotscalehigh}{HTML}{B54420}
\pgfplotsset{
  compat=1.16,
  paper curve/.style={densely dotted},
  paper corrected/.style={paper curve,plotcorrected,mark=square*,mark size=1.4pt,mark options={solid}},
  paper uncorrected/.style={paper curve,plotuncorrected,mark=triangle*,mark size=1.8pt,mark options={solid}},
  paper unenriched/.style={paper curve,plotunenriched,mark=o,mark size=1.5pt,mark options={solid,fill=white,line width=0.65pt}},
  paper initial/.style={paper curve,black!70,line width=0.9pt,mark=x,mark size=1.8pt,mark options={solid,line width=0.8pt}},
  paper axis/.style={
    font=\footnotesize,
    label style={font=\footnotesize},
    tick label style={font=\footnotesize},
    title style={font=\footnotesize},
    legend style={font=\footnotesize,draw=none,fill=none,cells={anchor=west}},
    axis line style={black!65},
    tick style={black!65},
    tick align=outside,
    grid=major,
    grid style={black!10},
    minor tick num=0,
    scaled ticks=false,
    /pgf/number format/1000 sep={},
    line width=0.85pt,
    mark size=2pt,
  },
  compact paper axis/.style={
    paper axis,
    tick label style={font=\scriptsize},
    legend style={font=\scriptsize,draw=none,fill=none,cells={anchor=west}},
    line width=0.8pt,
    mark size=1.5pt,
  },
}

\title{Understanding Enrichment in Reinforcement Learning}

\author{Jinwoo Kim
\thanks{Work done while an intern at Microsoft Research.} \\
University of California-San Diego\\
La Jolla, CA, USA \\
\texttt{jik083@ucsd.edu} \\
\And
Shraddha Barke \\
Microsoft Research \\
Redmond, WA, USA \\
\texttt{sbarke@microsoft.com}
}

\iclrfinalcopy 
\begin{document}

\maketitle
\lhead{Preprint}

\begin{abstract}
When rewards are sparse, reinforcement learning with verifiable rewards (RLVR) 
often uses hints or intermediate guidance to generate more successful rollouts.
This enrichment biases policy-gradient updates unless corrected via importance weights, 
but existing methods omit correction or truncate importance weights 
in order to avoid the high variance of correction.
It thus remains unclear what exactly is gained or lost in RLVR by correcting enriched rollouts.
We show, mathematically, that omitted or truncated correction implicitly reweights the defined reward, 
and we decompose the resulting gradient error into scale, rotation, and variance to explain their 
distinct effects on learning.
To make correction practical, we develop a novel sequential Monte Carlo (SMC) 
weight correction mechanism that, under stability and particle-order assumptions,
tempers the exponential compounding of standard correction variance over the length of a sample
to an additive accumulation.
We then apply our analysis of enrichment to interpreting the results of a fine-tune of 
Qwen3-1.7B on
a sparse band of OpenMathReasoning, 
establishing a concrete mechanism of how enrichment, 
both corrected and uncorrected,
helps avoid collapse in sparse domains.
Our main contribution is to understand, in general, how enrichment and correction can affect training, 
as opposed to claiming that either mode of operation is superior to unenriched RL.
\end{abstract}

\section{Introduction}
\label{sec:1intro}

Reinforcement learning with verifiable rewards (RLVR) has 
produced strong results in a wide variety of domains~\citep{lambert2024tulu,guo2025deepseek,wei2026swe,ren2025deepseek,jin2025search},
its primary advantage being that the model can learn from its own generations and 
a deterministic verifier.
However, RLVR struggles when the policy initially has a low success rate:
verifier rewards are often binary 
(e.g., 1 if a math problem is solved correctly, 0 if not), 
and thus an incompetent policy 
often ends up generating many zero-reward samples (or entire groups)
offering no learning signal~\citep{zhang2026scaf}.
Failure to generate a learning signal forces more rollouts 
during training, making RLVR inefficient in some cases~\citep{le2026no,foster2025learning}.

A common response to this problem is to \emph{enrich} the policy with
external guidance that makes successful trajectories more probable. 
Several methods supply hints or solution prefixes and generate rollouts under the 
hinted context~\citep{agrawal2026off,zhang2026bread,huang2025blending,zhang2025adhint,li2026questa}.
Others use process feedback to guide tree-structured rollouts:
PaTR~\citep{li2026process}, for example, uses process scorers,
including pretrained process reward models (PRMs), to expand promising
partial trajectories and stop unpromising branches early, then trains
on the resulting rollout groups with GRPO using task-outcome rewards.


A key observation is that such enriched policies \emph{alter the distribution} 
from which samples are drawn.
Calling the non-enriched base policy $\pi_{\theta}$ for a set of parameters 
$\theta$ (e.g., the parameters of an LLM), the goal of reinforcement learning 
(RL) is to maximize the expectation of a reward function $R$ on samples $x$ 
drawn from $\pi_{\theta}$, whose gradient is:
{\small
\begin{equation}
g =  \nabla_{\theta} J(\theta)
   = \mathbb{E}_{x\sim\pi_\theta}
     [R(x)\nabla_\theta\log\pi_\theta(x)]
\label{eq:intro-true-gradient}
\end{equation}
}

Computing a gradient $u$ 
over samples from an enriched policy $q_{\theta}$ instead of 
$\pi_\theta$ changes its value:
{\small
\begin{equation}
u = \mathbb{E}_{x\sim q_\theta}
     [R(x)\nabla_\theta\log\pi_\theta(x)]
\label{eq:intro-uncorrected-gradient}
\end{equation}
}
Clearly $u \neq g$ in general, so updating $\theta$
with the enriched gradient $u$ can lead to a different training outcome.
It is possible to \emph{correct} $u$, by weighing the contribution of each sample 
to counterbalance the difference in sampling rate:
computing $\mathbb{E}_{x\sim q_\theta}
     [\frac{\pi_{\theta}(x)}{q_{\theta}(x)} \cdot R(x)\nabla_\theta\log\pi_\theta(x)]$
instead of \Cref{eq:intro-uncorrected-gradient}.

However, many existing enrichment-based RLVR
methods~\citep{li2026questa,zhang2026bread,zhang2025adhint} 
do not compute the correction, 
or compute only a clipped approximation~\citep{agrawal2026off}. 
Two reasons 
help explain this practice.
First, full correction 
obtained by multiplying the correction weight $\bpolicy(\sample) / \qpolicy(\sample)$ 
often 
results in significant variance when using a finite number of 
samples to estimate $g$ during training, 
especially for long-horizon samples generated using 
LLMs~\citep{vespo}.
Second, perhaps more importantly, existing work has shown that uncorrected
or partially corrected gradients, 
where the correction weight ${\pi_{\theta}(x)}/{q_{\theta}(x)}$ is truncated 
to stay within a fixed range, still often lead to good results.
Consequently, most existing work focuses on developing methods of enrichment whose 
benefits hold under uncorrected or partially corrected gradients, as opposed to 
directly investigating the effects of weight correction on training or 
methods to reduce variance correction.
This leads to the central question we aim to answer in this paper:

{\centering
\textit{What does enrichment do to RL training, and what is gained or lost by correcting for it?}
\par}

Our paper primarily investigates this question through a 
mathematical analysis 
that characterizes enrichment as \emph{implicitly modifying the objective}
used by RL, 
and correction as a process that recovers the declared objective
(\Cref{sec:reward-reweighting}).
We also observe that training on the modified objective does not necessarily always 
lead to a worse policy.
The reweighted rewards manifest in training as a bias in the expected gradient, 
and we decompose the bias into scale and rotation terms, along with an additional variance term
induced by finite estimation and correction, 
and isolate the effect that each element has on training through controlled 
experiments~(\Cref{sec:scale-effect,sec:rotation-effect}).

Our theory suggests that full correction may be worth investigating as an alternative for 
uncorrected enrichment in RL, but we observe that in LLMs-as-policies, the 
variance of the correction weights can be so high that standard correction results 
in computations that may not be stably representable in float32~\citep{vespo,sheng2025hybridflow}.
We thus develop a method for drawing rollouts from the enriched policy $\qpolicy$
while keeping the variance cost 
of correction low, based on sequential Monte-Carlo sampling (SMC)~\citep{smcbook}.
A key feature of our SMC-correction is an \emph{additive-in-length} variance bound 
(under 
some mathematical assumptions in \Cref{app:smc-additive-variance})
in contrast to the exponential bound of standard weight correction, 
and to the best of our knowledge, is the first approach that enables
practical, untruncated weight correction for real RL training.

Finally, we fine-tune Qwen3-1.7B~\citep{qwen3} on a sparse-success subset 
of OpenMathReasoning~\citep{moshkov2025aimo}, via
\rone uncorrected enriched, 
\rtwo SMC-corrected enriched, 
and \rthree unenriched RL 
as an application of our theory to analyze a real RL run (\Cref{sec:results}).
Our results identify a concrete mode of training where enrichment, 
including the uncorrected variant, becomes useful: when the base policy 
has a behavioral profile that has a small number of successes concentrated 
on a few problems, 
unenriched base RL collapses towards them.
Enrichment can expose the policy towards a more diverse set of successes and 
delay the collapse or avoid it altogether.

Overall, our paper is positioned as a study of enrichment 
and its correction, 
providing a theoretical basis for understanding enrichment 
and correction
that future work can build upon
rather than a novel algorithm that achieves state-of-the-art results.

\section{How Enriched Sampling Changes Learning}
\label{sec:2analysis}

We now investigate how enrichment 
impacts learning from a mathematical perspective.
Intuitively, uncorrected enrichment, in the ideal case, 
improves learning by implicitly \emph{amplifying} the reward assigned to samples 
that occur more frequently in the enriched policy compared to the base policy.
Operationally, this difference manifests itself as a bias on the 
expected gradient, which can further be decomposed into two main elements:
\rone \emph{scale} on the size of the gradient and 
\rtwo \emph{rotation} on the direction, 
with an additional error term introduced due to 
\rthree \emph{variance} of the estimator.

\textbf{Setting and notation.}
In this paper, we use 
$\bpolicy$ to denote the base policy (without enrichment) and 
$\qpolicy$ to denote the enriched policy, where
$\param$ refers to the shared set of parameters that define $\bpolicy$ and $\qpolicy$.
We use $\sample$ to denote a complete response from a policy $\policy$,
$\policy(\sample)$ to denote the probability of drawing $\sample$ from $\policy$, 
and $\reward(\sample)$ to denote the reward of $\sample$.
We focus on LLMs-as-policies where a sample $x$ is a series of tokens 
$x = t_1, \cdots, t_l$, abbreviated as $t_{1:l}$,
and a policy $\policy$ computes
next-token probabilities given a prefix (i.e., $\policy(t_i \mid t_{1:i-1})$).
We focus in this section on standard REINFORCE learning~\citep{williams1992simple}; 
an extension to
GRPO~\citep{shao2024deepseekmath} can be found in~\Cref{app:grpo-distortion}.


\subsection{Explaining enrichment as a redistribution of reward}
\label{sec:reward-reweighting}
Recall Equations
\ref{eq:intro-true-gradient} and \ref{eq:intro-uncorrected-gradient}, 
which 
compute the precise gradients when the sample is drawn from 
the base policy $\bpolicy$ and the enriched policy $\qpolicy$:
{\small
\begin{equation*}
g = \mathbb{E}_{x\sim \pi_\theta}
     [R(x)\nabla_\theta\log\pi_\theta(x)]
\quad (\ref{eq:intro-true-gradient})
\qquad \qquad 
u = \mathbb{E}_{x\sim q_\theta}
     [R(x)\nabla_\theta\log\pi_\theta(x)]
\quad (\ref{eq:intro-uncorrected-gradient})
\end{equation*}
}
In practice, $g$ and $u$ are \emph{estimated} via a set of $N$ independent samples 
$x_1, \cdots, x_N$ drawn from their respective policies.
Calling the policy in question $\policy$, in both cases the inner term 
$\reward (\sample) \nabla_\theta \log \bpolicy (\sample)$ expands to produce the 
following estimation over the $N$ samples:
{\small
\begin{equation}
\hphantom{\qquad\qquad (\sample_i \overset{\mathrm{i.i.d.}}{\sim} \policy).}
\hat{h} = \frac{1}{N} \cdot 
            \sum_{i=1}^{N} \frac{\nabla_\theta \bpolicy (\sample_i)}{\bpolicy (\sample_i)} \cdot 
            \reward (\sample_i)
\qquad\qquad (\sample_i {\sim} \policy).
\label{eq:uncorrected-gradient-estimator}
\end{equation}
}
The mismatch comes from the denominator $\bpolicy (\sample_i)$ and the 
distribution from which $\sample_i$ is drawn from.
If $\sample_i \sim \bpolicy$, then $\bpolicy (\sample_i)$ cancels out in 
expectation, and the mean of the unenriched estimator
$\hat{g}$ becomes $\sum_{\sample}\nabla_\theta \bpolicy (\sample) \cdot R(\sample)$: 
intuitively, to move the parameters $\param$ in a direction such that 
the probability ($\bpolicy (\sample)$) of drawing a rewarding sample 
(e.g., $\reward(\sample) = 1$ for binary rewards) is increased.

On the other hand, if $\sample_i \sim \qpolicy$, the cancellation does not happen 
and the intuitive reading of the gradient no longer holds.
In expectation, the estimator of the uncorrected gradient $\hat{u}$ 
instead becomes 
$\sum_{\sample}\frac{\nabla_\param \bpolicy (\sample)}{\bpolicy (\sample)} \cdot R(\sample) \cdot \qpolicy(\sample) = 
\sum_{\sample}\nabla_\param \bpolicy (\sample) \cdot \frac{\qpolicy(\sample)}{\bpolicy (\sample)}\; R(\sample)
$: 
the gradient points in a direction that increases the base policy $\bpolicy$'s 
expected \emph{distorted} reward 
$\rreward(\sample) = \frac{\qpolicy (\sample)}{\bpolicy (\sample)} \reward(\sample)$, 
instead of the defined reward $\reward(\sample)$.
We thus observe that the effect of enrichment can be explained as an 
implicit \emph{reweighting} of the reward $\reward$, where the reward of a sample $\sample$
is \emph{amplified} if it occurs more frequently in the enriched policy $\qpolicy$ compared to the 
base policy $\bpolicy$, 
and \emph{attenuated} in the reverse case.

\begin{example}[Unequal reinforcement of successful responses]
\label{ex:unequal-reinforcement}
Consider an example task with two correct responses 
$x_a,x_b$ and one incorrect response $x_f$, 
whose rewards are $1,1,0$.
Suppose the base policy $\bpolicy$ assigns them probabilities 
$(0.01,0.04,0.95)$, while the enriched policy $\qpolicy$ assigns 
the probabilities $(0.10,0.02,0.88)$.
The effective rewards are then
{\small
\[
  \rreward (x_a)=\frac{0.10}{0.01}\cdot1=10,
  \qquad \rreward (x_b)=\frac{0.02}{0.04}\cdot1=\frac12,
  \qquad \rreward (x_f)=0.
\]
}
Enrichment raises the overall reward hit rate from $5\%$ to $12\%$, 
but also reweights the effective rewards of $\sample_a$ and $\sample_b$: 
$\sample_a$ now has a reward 20 times larger than that of $\sample_b$.
Compared to the unenriched gradient, 
enrichment amplifies the contribution that pushes toward $\sample_a$ 
tenfold while halving the contribution that pushes toward $\sample_b$, 
despite both responses receiving the same verifier reward.
\end{example}

This reweighting can \emph{help} 
when enrichment assigns high probability to successful responses that the base policy 
rarely generates, increasing their training signal;
i.e., the reweighting has the effect of teaching the model
the successes of the enriched policy 
as opposed to reinforcing the base policy's own successes
(we will see a case of this phenomenon in \Cref{sec:results}).
Conversely, enrichment that recognizes only a familiar solution family can disproportionately 
reinforce the family and steer learning away from alternatives, 
including higher-reward solutions as illustrated in \Cref{sec:rotation-effect}.

\subsection{Decomposing Gradient Error into Scale, Rotation, and Variance}
\label{sec:gradient-decomposition}

We observe that the modified reward $\rreward$ 
in \Cref{sec:reward-reweighting}
is dependent on the parameters $\theta$, 
and thus one cannot 
simply say 
uncorrected enriched training is equivalent to REINFORCE training under 
a different fixed reward: the reweighted reward changes each step.
Thus the best way to understand the impact of implicit reweighting is through the possible 
gradient distortions reweighting gives rise to at each update during training: 
an effect that can be decomposed into scale, rotation, and variance.


Let $g \neq 0$ denote the true gradient defined in \Cref{eq:intro-true-gradient}, 
$u$ an uncorrected gradient,
and $\hat{u}$ an estimator of $u$ such that the mean of the estimator $\e[\hat{u}] = u$.
We observe that $\hat{u}$ admits a standard mean-squared-error decomposition 
against $g$
with 
bias 
and variance terms. 
The bias term can be further split into parallel 
(a \emph{scale} $\alpha$ on $g$) and perpendicular
(a \emph{rotation} $r$ relative to $g$) components:
{\small
\begin{equation}
  \mathbb{E}\|\widehat u-g\|^2
  =\underbrace{(\alpha-1)^2\|g\|^2}_{\text{scale error}}
   +\underbrace{\|r\|^2}_{\text{rotation error}}
   +\underbrace{\mathbb{E}\|\widehat u-u\|^2}_{\text{sampling variance}}.
  \label{eq:three-way-mse}
\end{equation}
}
Here $\alpha=\langle u,g\rangle/\|g\|^2$ is the signed scale factor and 
$r=u-\alpha g$ is perpendicular to $g$.
Each term in \Cref{eq:three-way-mse} has a different consequence for training, 
which we isolate and illustrate the effect of via controlled experiments, below.


\subsubsection{Scale: amplifying the learning signal}
\label{sec:scale-effect}

A \emph{scale} factor arises when the enrichment $\qpolicy$ consistently prefers samples $\sample$ 
where 
the reward 
$\reward(\sample) > 0$ by a factor of $\alpha$: i.e., 
$\forall \sample, \qpolicy (\sample) = \alpha  \bpolicy (\sample)$.
The derivation in \Cref{sec:reward-reweighting} shows that such consistent oversampling results in 
an amplified (or attenuated, if $\alpha < 1$) reward $\rreward = \alpha \reward$.
If training is unregularized, in expectation, a scale factor 
does not affect training outside of 
an increase in how quickly training proceeds and its stability depending on step size.

The interesting difference comes when training is \emph{regularized}:
for example, 
by a KL-penalty term $K(\param)$, 
such that the objective $J(\param)$ becomes 
$\e_{x \sim \bpolicy}[\reward (\sample)] - \lambda \cdot K(\param)$ 
instead of the unregularized expected reward $\e_{x \sim \bpolicy}[\reward (\sample)]$.
Uncorrected enrichment amplifies the reward in $J(\param)$
but not the KL-penalty term, and the
mean of the uncorrected estimator $\e[\hat{u}]$ becomes
in expectation:
{\small
\begin{equation}
  \e[\hat{u}] = \alpha g-\lambda\nabla_\param K(\param)
  =\alpha\left(g-\frac{\lambda}{\alpha}\nabla_\param K(\param) \right).
  \label{eq:effective-regularization}
\end{equation}
}
In other words, the strength of the KL-penalty term is altered by a factor of $1 / \alpha$ after 
normalization.
We observe that 
unlike the unregularized scenarios,
this attenuation cannot be recovered by changing other hyperparameters, 
such as learning rate: it represents a fundamental change to the objective.

\textbf{Training Illustration.}
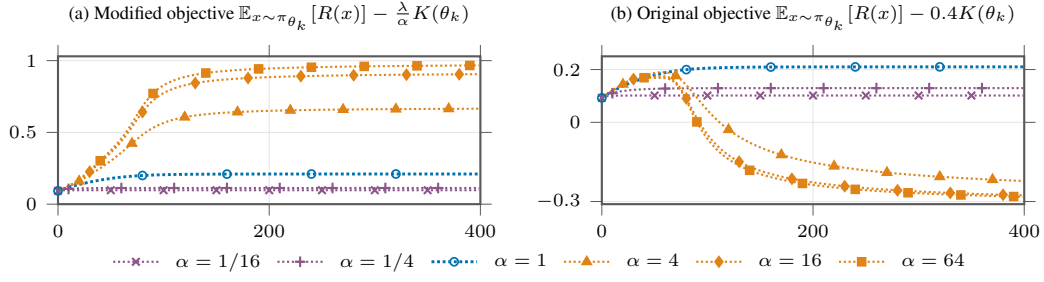
\begin{figure}[!tb]
  \centering
\begingroup
\pgfplotsset{
  scale attenuated sixteen/.style={paper uncorrected,plotscalelow,mark=x,mark size=2pt,mark options={solid,line width=0.8pt},mark repeat=50,mark phase=1},
  scale attenuated four/.style={paper uncorrected,plotscalelow,mark=+,mark size=2pt,mark options={solid,line width=0.8pt},mark repeat=50,mark phase=11},
  scale baseline/.style={paper unenriched,line width=1.1pt,mark repeat=80},
  scale amplified four/.style={paper uncorrected,mark repeat=50,mark phase=21},
  scale amplified sixteen/.style={paper uncorrected,mark=diamond*,mark repeat=50,mark phase=31},
  scale amplified sixtyfour/.style={paper uncorrected,mark=square*,mark size=1.4pt,mark repeat=50,mark phase=41},
}
\begin{tikzpicture}
\begin{groupplot}[
  compact paper axis,
  /tikz/font=\scriptsize,
  label style={font=\scriptsize},
  tick label style={font=\scriptsize},
  title style={font=\scriptsize,align=center},
  group style={group size=2 by 1,horizontal sep=0.115\linewidth},
  width=0.40\linewidth,height=0.14\linewidth,scale only axis,
  xmin=0,xmax=400,xtick={0,200,400},
]
\nextgroupplot[
  title={(a) Modified objective $\mathbb{E}_{x\sim\pi_{\theta_k}}[R(x)]-\frac{\lambda}{\alpha}K(\theta_k)$},
  ymin=0,ymax=1.03,ytick={0,0.5,1},
  legend columns=6,
  legend style={font=\scriptsize,at={(1.14375,-0.25)},anchor=north,column sep=4pt},
]
\addplot[scale attenuated sixteen] table[x=step,y=modified_objective_a16inv,col sep=comma]{figs/data/scale.csv};
\addlegendentry{$\alpha=1/16$}
\addplot[scale attenuated four] table[x=step,y=modified_objective_a4inv,col sep=comma]{figs/data/scale.csv};
\addlegendentry{$\alpha=1/4$}
\addplot[scale baseline] table[x=step,y=modified_objective_a1,col sep=comma]{figs/data/scale.csv};
\addlegendentry{$\alpha=1$}
\addplot[scale amplified four] table[x=step,y=modified_objective_a4,col sep=comma]{figs/data/scale.csv};
\addlegendentry{$\alpha=4$}
\addplot[scale amplified sixteen] table[x=step,y=modified_objective_a16,col sep=comma]{figs/data/scale.csv};
\addlegendentry{$\alpha=16$}
\addplot[scale amplified sixtyfour] table[x=step,y=modified_objective_a64,col sep=comma]{figs/data/scale.csv};
\addlegendentry{$\alpha=64$}
\nextgroupplot[
  title={(b) Original objective 
  $\vphantom{\frac{\lambda}{\alpha}}\mathbb{E}_{x\sim\pi_{\theta_k}}[R(x)]-0.4K(\theta_k)$},
  ymin=-0.31,ymax=0.25,ytick={-0.3,0,0.2},
]
\addplot[scale attenuated sixteen] table[x=step,y=F_a16inv,col sep=comma]{figs/data/scale.csv};
\addplot[scale attenuated four] table[x=step,y=F_a4inv,col sep=comma]{figs/data/scale.csv};
\addplot[scale baseline] table[x=step,y=F_a1,col sep=comma]{figs/data/scale.csv};
\addplot[scale amplified four] table[x=step,y=F_a4,col sep=comma]{figs/data/scale.csv};
\addplot[scale amplified sixteen] table[x=step,y=F_a16,col sep=comma]{figs/data/scale.csv};
\addplot[scale amplified sixtyfour] table[x=step,y=F_a64,col sep=comma]{figs/data/scale.csv};
\end{groupplot}
\end{tikzpicture}
\endgroup
  \caption{
  Scale-modified (left, y-axis) and original objectives (right, y-axis) tracked across training steps $k$
  (x-axis), for the scale-isolated simulated training runs, over different values of the scale 
  coefficient $\alpha$.
  We use $\theta_k$ to denote the parameters at step $k$.
  }
  \label{fig:scale-effect}
\end{figure}
To clarify the effect this modification has on training, we
simulate a training run over an objective 
$J(\param) = \e_{x \sim \bpolicy}[\reward (\sample)] - \lambda \cdot K(\param)$, 
where the task is simple enough we can compute precise gradients for 
for the reward objective 
$\nabla_\param \e_{\sample \sim \bpolicy}[\reward (\sample)]$ 
and the KL-penalty
$\nabla_\param \lambda K(\param)$
directly (without sampling) ($K(\theta) = \kld(\pi_\theta\|\pi_{\theta_0})$ and 
$\param_0$ denotes the initial policy parameters).
We compute the gradient a scale factor would induce by applying the $1/\alpha$ 
multiplication to the 
KL-gradient and simulate training by applying the gradient update over multiple timesteps 
while varying $\alpha$ from $1/16$ to $64$; details are in 
\Cref{app:illustrative-setup}.


\Cref{fig:scale-effect} tracks how the modified (panel (a)) 
and the original (panel (b)) objectives evolve throughout training.
The contrast illustrates the effect of scale distortion: the modified objective 
rises as training proceeds, but for $\alpha > 1$, the original objective \emph{falls} after 
peaking at around step 70 as the model implicitly optimizes for an attenuated KL penalty.
For $\alpha < 1$ the original objective does not fall, 
but training plateaus at a much
lower reward, being constrained by a much stronger KL penalty.
We observe that for the $\alpha > 1$ runs in particular, a monitor tracking 
the modified objective (e.g., by reusing training samples from $\qpolicy$) 
could suggest 
that training is progressing successfully and 
hide that the true objective is decreasing silently.
On the other hand, we also note that deviation from the original objective 
does not always mean bad performance: the $\alpha > 1$ cases did achieve a higher raw reward, 
which may be a desirable tradeoff against the 
increased KL-deviation in 
real-world scenarios.

\subsubsection{Rotation: changing which solutions are reinforced}
\label{sec:rotation-effect}

The scale-only case requires amplifying all nonzero rewards equally.
In practice, enrichment will likely prefer a 
certain family of solutions over another~\citep{shen2026boosting};
the reweighting in \Cref{sec:reward-reweighting}
shows this leads to the reward of preferred samples being \emph{relatively}
amplified compared to those that are not.
This relativity manifests as a \emph{rotation} in the gradient space:
rotated gradients follow different trajectories and can become trapped
in (or escape from) different suboptimal policies.

\textbf{Training Illustration.}
We illustrate the local minima effect using a similar training setup as in 
\Cref{sec:scale-effect}, 
where this time, the 
objective is unregularized and tracks raw expected reward 
$\e_{\sample \sim \bpolicy}[\reward (\sample)]$.
In addition to the base policy $\bpolicy$ being trained, we construct an enriched policy 
$\qpolicy$ from $\param$ that prefers $0.5$-reward samples but avoids $1$-reward samples.
We then compute the precise gradients 
$g$ and $u$ exactly, 
normalize both gradients to unit Euclidean norm,
and perform a simulated training run as in \Cref{sec:scale-effect}.
The training settings create an initial angle between $g$ and $u$ of roughly $39$ degrees, 
eventually falling to roughly $7$ degrees.
Details can be found in \Cref{app:illustrative-setup}.

\begingroup
\setlength{\intextsep}{4pt}
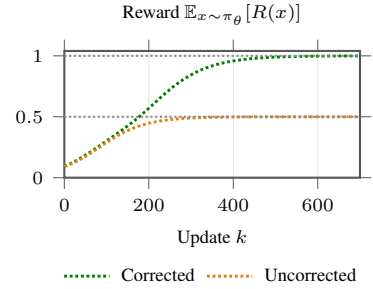
\begin{wrapfigure}{r}{0.40\textwidth}
  \centering
  \begin{tikzpicture}
\begin{axis}[
  compact paper axis,
  /tikz/font=\scriptsize,
  label style={font=\scriptsize},
  width=0.70\linewidth,height=0.30\linewidth,scale only axis,
  ymin=0,ymax=1.04,ytick={0,0.5,1},
  title={Reward $\e_{x \sim \bpolicy}[\reward (\sample)]$},
  title style={font=\scriptsize},xlabel={Update $k$},
  xmin=0,xmax=700,xtick={0,200,400,600},
  legend columns=2,
  legend style={at={(0.5,-0.64)},anchor=north,column sep=2pt,inner xsep=0pt,inner ysep=1pt},
]
\addplot[paper corrected,no marks,line width=1.1pt] table[x=step,y=J,col sep=comma]{figs/data/rotation_g.csv};
\addlegendentry{Corrected}
\addplot[paper uncorrected,no marks,line width=1.1pt] table[x=step,y=J,col sep=comma]{figs/data/rotation_u.csv};
\addlegendentry{Uncorrected}
\addplot[black!45,densely dotted,forget plot] coordinates {(0,0.5) (700,0.5)};
\addplot[black!45,densely dotted,forget plot] coordinates {(0,1) (700,1)};
\end{axis}
\end{tikzpicture}
  \small
  \setlength{\abovecaptionskip}{3pt}
  \caption{
    Reward evolution across update steps for uncorrected (orange) and
    corrected enriched \mbox{(green) training.}
  }
  \label{fig:rotation-effect}
\end{wrapfigure}

\Cref{fig:rotation-effect} tracks the true objective
$\e_{\sample \sim \bpolicy}[\reward(\sample)]$
across training using the unit-normalized corrected ($g$) and uncorrected ($u$) gradients.
With update lengths equalized, uncorrected training plateaus
at reward $0.5$, producing the $0.5$-reward strings preferred by enrichment.
Unenriched training, on the other hand, eventually reaches a policy $1$-reward 
samples.
The relative amplification explained in this section predicts exactly this phenomenon: 
in our setup, the enriched policy $\qpolicy$ prefers the $0.5$-reward samples by 
4.3 times more than the $1$-reward samples. 
This preference translates
into an effective reward of 2.15 for $0.5$-reward samples versus 0.46 for $1$-reward samples 
steering uncorrected training towards the suboptimal $0.5$-samples instead.
This phenomenon is \emph{not} a local minima that can be easily escaped; 
if, e.g., $\qpolicy$ consistently prefers the $0.5$-reward samples 
throughout the training trajectory, then the $0.5$-sample rewards will always be amplified 
and training will consistently pull towards these samples.
On the other hand, one can also see how an enrichment rotated towards high-reward samples 
would be highly useful in avoiding low-reward local minima during training, by amplifying their reward.
\par
\endgroup


\subsubsection{Variance: the cost of restoring the intended update}
\label{sec:variance-effect}

The illustrative experiments in 
\Cref{sec:scale-effect,sec:rotation-effect} give concrete examples of how 
training can be negatively affected due to uncorrected gradients.
Although they do \emph{not guarantee} suboptimality (indeed, distortion from enrichment can be 
beneficial as we will see in \Cref{sec:results}),
it stands to reason that enriched, corrected gradients are also worth consideration.
However, as discussed in \Cref{sec:1intro}, standard weight correction comes at a high 
cost of \emph{variance} 
when estimating $g$ with a corrected finite estimator $\hat{g}$, 
which can have its own undesirable 
consequences for training.
The key behavior is when a rewarding sample $x$ is \emph{undersampled} in the 
enriched policy $\qpolicy$, compared to the base policy $\bpolicy$ (i.e., 
the correction weight $(\bpolicy (\sample)) / \qpolicy (\sample) > 1$).
When this ratio is large, 
a finite batch during training may either 
\rone miss $x$ altogether, reinforcing other rewarding samples preferred by $\qpolicy$, or
\rtwo the large corrected weight of $x$ can dominate the batch 
and produce an abrupt parameter change; 
training can therefore depend strongly on if and when these two cases happen.

To make these two cases concrete, 
fix one rewarding response $\sample$ that has a small but positive probability under $\qpolicy$, 
and let $\hat{g}$ average the importance-corrected contributions from $N$ independent samples 
$\sample_1,\ldots,\sample_N\sim\qpolicy$.
If the batch does not contain $\sample$, it receives none of the gradient contribution associated with 
$x$; this happens with probability
  $\p(\forall i, 1 \leq i \leq N. \ \sample_i\ne\sample)
  =\bigl(1-\qpolicy(\sample)\bigr)^N$;
when $N \cdot \qpolicy(\sample)\ll1$, most batches are of this kind.

On the other hand, a single occurrence of $\sample$ in the batch contributes
{\small
\begin{equation}
  \frac{1}{N}\frac{\bpolicy(\sample)}{\qpolicy(\sample)}
  \reward(\sample)\nabla_\param\log\bpolicy(\sample)
  \label{eq:rare-sample-contribution}
\end{equation}
}
\noindent
To the estimated gradient $\hat{g}$ ($k$ occurrences in a batch contribute this term $k$ times).
In expectation, the contribution in \Cref{eq:rare-sample-contribution} 
averages to the desired
$\bpolicy(\sample)\reward(\sample)\nabla_\param\log\bpolicy(\sample)$, but the 
discrepancy lies in the fact that a single occurrence of $x$
supplies
$1/(N \cdot \qpolicy(\sample))$ times this average contribution to the gradient.
Thus when $N \cdot \qpolicy(\sample) \ll 1$
(which LLMs-as-policies are particularly prone to), 
the contribution becomes very large and the 
gradient can also entail an abrupt parameter change dominated by $x$.
We observe that it is not just high variance in the gradients, but this particular bimodal manifestation, that becomes undesirable for training and is thus avoided in the 
literature~\citep{li2026questa,zhang2026bread,zhang2026scaf,zhang2025adhint}
under the umbrella term `variance'.

\textbf{Training Illustration.}
We illustrate the effects of variance through another controlled experiment 
where the enriched policy 
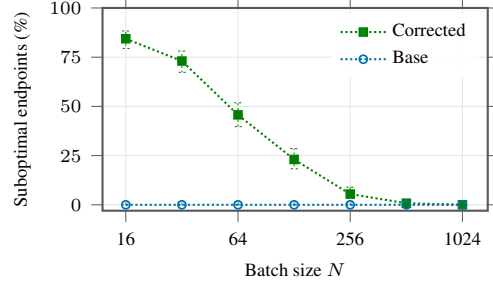
\begin{wrapfigure}{r}{0.48\textwidth}
  \centering
\begin{tikzpicture}
\begin{axis}[
  compact paper axis,
  width=0.76\linewidth,height=0.40\linewidth,scale only axis,
  /tikz/font=\scriptsize,
  label style={font=\scriptsize},
  xlabel={Batch size $N$},
  ylabel={Suboptimal endpoints (\%)},
  xmode=log,log basis x=2,
  xmin=12,xmax=1400,xtick={16,64,256,1024},
  xticklabels={16,64,256,1024},
  ymin=-3,ymax=100,ytick={0,25,50,75,100},
  legend style={at={(0.98,0.98)},anchor=north east,fill=white,fill opacity=0.9,text opacity=1},
]
\addplot[paper corrected,error bars/.cd,y dir=both,y explicit]
  table[x=N,y=corrected,y error minus=corrected_minus,y error plus=corrected_plus,col sep=comma]{figs/data/variance_batch.csv};
\addlegendentry{Corrected}
\addplot[paper unenriched,error bars/.cd,y dir=both,y explicit]
  table[x=N,y=iid,y error minus=iid_minus,y error plus=iid_plus,col sep=comma]{figs/data/variance_batch.csv};
\addlegendentry{Base}
\end{axis}
\end{tikzpicture}
  \footnotesize
  \caption{Percentage of reward $<0.9$ endpoints 
  generated by  
  \rone corrected enriched and
  \rtwo base
  REINFORCE, 
  mapped against batch size.
  }
  \label{fig:variance-effect}
\end{wrapfigure}
$\qpolicy$ prefers $0.5$-reward samples strongly compared to 
$1$-reward samples, as in \Cref{sec:rotation-effect}.
We start from an initial base policy $\bpolicy$ that generates $1$-reward samples 
with probability $\approx 0.6$; $\qpolicy$ instead prefers $0.5$-reward samples
and gives $1$-reward samples only $\approx 3.0 \cdot 10^{-5}$ probability: 
i.e., $\qpolicy$ severely undersamples $1$-reward samples.

We perform two types of training: base REINFORCE using samples from $\bpolicy$, and 
\emph{corrected} 
(via weight ratios)
enriched REINFORCE using samples from $\qpolicy$: the correction makes the expected 
update equivalent for both approaches, but only the latter configuration shows the bimodal 
variance we discussed.
We vary the batch size $N$ and perform 256 training runs for each configuration 
$(N, \mathrm{policy})$, measuring the number of runs failing to reach
a policy where reward $\geq 0.9$.

\Cref{fig:variance-effect} illustrates this percentage mapped across batch sizes:
the graph shows that when batch size $N=16$, over 80\% of policies end with reward $< 0.9$ 
under the corrected configuration.
This effect is alleviated as $N$ (and $N \cdot \qpolicy(\sample)$) grows larger, 
but still visible at $N=256$ where $5.5\%$ of runs finish with reward $< 0.9$.
Base REINFORCE avoids this problem; all runs succeed in reaching policies 
with reward $\geq 0.9$ regardless of batch size, \emph{even though} the expected update for 
both approaches are equivalent.
This shows precisely how bimodal variance from weight correction can be harmful, 
and why existing work~\citep{li2026questa,zhang2026bread,zhang2026scaf,zhang2025adhint} refrains from applying correction.

\section{Correcting Enriched Rollouts with Sequential Monte Carlo}
\label{sec:smc}

\Cref{sec:variance-effect} illustrates why bimodal variance can make 
weight correction undesirable for RL.
In addition, we observe that for modern LLMs-as-policies, accumulated weights can be so large 
that computations on them cannot be represented stably in float32, 
making weight correction impractical.

The main contributor is that 
the correction weight 
$\bpolicy (\sample) / \qpolicy (\sample)$ accumulates \emph{multiplicatively} as 
the length of the sample increases.
For $\sample = t_1 \cdots t_l$, we have that 
$\frac{\bpolicy (\sample)}{\qpolicy (\sample)} = 
\prod_{j = 1}^{l} \frac{\bpolicy(t_j \mid t_{1:j-1})}{\qpolicy(t_j \mid t_{1:j-1})}$:
i.e.,
the total correction weight is the \emph{product} of the per-token weight ratios returned by the LLM 
at each position.
If the enriched policy consistently over- or undersamples the sample $x$ at token-level granularity
(which is relatively common in real LLM policies~\citep{zhang2026rethinking}), 
the total correction weight becomes exponential in the length of the sample.
Verl~\citep{sheng2025hybridflow}, an RL framework, thus internally enforces an log-upper bound 
of $20$
on the size of the correction weight 
as a conservative guard to ensure float-32 representability for downstream computations.
Our preliminary experiments from \Cref{sec:results} often triggered this bound, indicating that 
naive weight correction is impractical, even if one is willing to accept the variance cost.

Thus to consider weight correction in any practical scenario, one must have some method 
of performing weight correction while keeping variance low.
This section satisfies this goal by 
formulating a variant of 
sequential Monte-Carlo sampling (SMC)~\citep{zhao2024probabilistic} 
to draw samples from the enriched policy $\qpolicy$ while reducing the variance cost of correction.
In particular, 
SMC-style resampling tempers the 
multiplicative accumulation of correction weights, 
with an additive-in-length variance bound under the stability and particle-order assumptions in \Cref{app:smc-additive-variance}.



\subsection{Reducing Gradient Variance via SMC Resampling}
\label{sec:smc-sequential}

Let us assume we are generating all $N$ samples for a batch-gradient in parallel lockstep 
(i.e., we draw the first token for all $N$ samples, then the second token for all $N$ samples, 
and so on instead of generating each sample sequentially).
For each sample $\sample_i$, during generation, we first attach $\sample_i$ with a 
\emph{correction weight} $\cweight_i$ that gets updated incrementally as each token gets drawn:
$\cweight_i = \cweight_i \cdot \frac{\bpolicy (t_k | t_{1:k-1})}{\qpolicy (t_k | t_{1:k-1})}$ 
and the initial $\cweight_i = 1$.
\footnote{
  We focus on per-token updates, but the idea transfers to coarser 
  generation boundaries as well.
}
We observe that generating with $\cweight_i$ attached without any intervention will eventually 
yield $\cweight_i = \bpolicy(\sample_i) / \qpolicy(\sample_i)$, the full correction weight, for each 
$\sample_i$.

A key intuition is that having incremental weights allows us to detect whether the variance 
in the final sample set is likely to become large \emph{during} generation: by tracking the variance 
between
$W_i$ 
(i.e., the effective sample size (ESS) in terms of SMC~\citep{smcbook}).
When this variance passes a certain threshold
we trigger a \emph{weighted resample} of the $N$ current partial samples, 
proportional to their correction weights $W_i$, then \emph{reset} each weight to the mean of $W_i$. 
\begin{example}[Variance Reduction]
  \label{ex:variance-reduction}
  Suppose $N=4$ partial samples $(x_1, x_2, x_3, x_4)$ have correction weights 
  $(W_1, W_2, W_3, W_4) = (3,\tfrac13,\tfrac13,\tfrac13)$, and one token remains to be generated.
  Let us assume that 
  there are two possible final tokens, $\texttt{t}$ and $\texttt{u}$, where at the final position
  their respective probabilities are $(1/4, 3/4)$ under $\bpolicy$ and 
  $(1/2, 1/2)$ under $\qpolicy$.
  Each final token therefore multiplies a correction ratio of either $1/2$ or $3/2$ 
  (with equal probability); the variance of the ratio is $1/4$.

  Without resampling, $W_1$ is either $3/2$ or $9/2$, while $W_2$, $W_3$, and $W_4$ are either 
  $1/6$ or $1/2$, all with equal probabilities.
  On the other hand, with resampling, $x_1$ gains three copies in expectation while 
  $x_2, x_3, x_4$ each gain $1/3$ each.
  The importance weights at this point are \emph{reset} to the mean weight 1: 
  thus after the final token is drawn, all 
  samples have $W_i = 1/2$ or $3/2$.
  We observe that the expected variance of the correction weights in each scenario is:
  {\small
  \begin{equation*}
  \begin{array}{@{}c@{\qquad}c@{}}
    \text{without resampling:} & \text{with resampling:} \\[0.5ex]
    \displaystyle
      \underbrace{\frac{1}{4^2}}_{\substack{\text{batch-mean}\\\text{variance factor}}}
      \underbrace{\left(3^2+3\left(\frac13\right)^2\right)}_{\substack{\text{sum of squared}\\\text{attached weights}}}
      \underbrace{\frac14}_{\substack{\text{variance of the}\\\text{last-token ratio}}}
      =\frac7{48}
    &\displaystyle
      \underbrace{\frac{1}{4^2}}_{\substack{\text{batch-mean}\\\text{variance factor}}}
      \underbrace{\left(1^2+1^2+1^2+1^2\right)}_{\substack{\text{sum of squared}\\\text{reset weights}}}
      \underbrace{\frac14}_{\substack{\text{variance of the}\\\text{last-token ratio}}}
      =\frac1{16}.
  \end{array}
  \end{equation*}
  }
  Resampling has reduced this variance by a factor of $7/3$, 
  while still preserving the expected contribution of each expected sample 
  (because high-$W_i$ samples
  generate more clones during resampling).
  The key reduction factor comes from the \emph{remaining} variance induced by completing the 
  partial samples.
  Without resampling, 
  $\sample_1$'s final weight depends on one final-token draw multiplied by $3$,
  while after resampling, 
  the contribution is spread across independent draws with weight 1, 
  reducing variance of the contribution: 
  this is the intuitive mechanism in how the SMC formulation 
  tempers multiplicative variance accumulation into additive accumulation.
\Cref{app:smc-experiment} provides a controlled experiment 
illustrating the variance reduction on gradients.

\end{example}

Once generation is complete, we use the final correction weights $\cweight_i$ to estimate the gradient 
with correction.
The estimator
$\hat{g}_{\mathrm{SMC}}$ is then defined as and satisfies:
{\small
\begin{equation}
  \hat{g}_{\mathrm{SMC}}
  =\frac1N\sum_{i=1}^{N}
    \cweight_i\reward(\sample_i)\nabla_\param\log\bpolicy(\sample_i),
  \qquad
  \e[\hat{g}_{\mathrm{SMC}}]=g.
  \label{eq:smc-basic-gradient}
\end{equation}
}
Resampling preserves the expected total correction weight of each prefix, 
which is the intuition in how SMC resampling preserves the mean of the expected gradient.
We also note that the resampling mechanism induces a variance overhead, 
due to the fact that resampling copies prefixes (effectively reducing 
the number of distinct samples).
Precise assumptions and a derivation of the additive-in-length bound,
variance overhead, 
as well as a proof 
that $\e[\hat{g}_{\mathrm{SMC}}]=g$, is provided in \Cref{app:smc-additive-variance}.
\Cref{app:smc-guidance} also provides an extension of the discussed SMC mechanism that 
that introduces a separate enrichment weight 
for better enrichment tracking.

One important qualification is that mean preservation 
as in \Cref{eq:smc-basic-gradient}
does not automatically 
extend to GRPO, which can
introduce both scale and rotation bias.
However, we note that this distortion is not unique to SMC correction: 
even unenriched GRPO can induce a scale bias on individual problem gradients, 
which can further aggregate into a rotation bias when gradients are 
aggregated across problems (we illustrate this in \Cref{app:grpo-distortion}).
\citet{yao2026group} provide a pairwise-surrogate justification of such group-relative updates; 
their weighted version applies to our own SMC-corrected reward update and establishes a 
principled practical use of SMC-correction within GRPO.

\section{Uncorrected and Corrected Enrichment in Practice}
\label{sec:results}
We now connect our analysis to a real training run, by performing three different modes of 
RL to train Qwen3-1.7B~\citep{yang2025qwen3}: 
\rone enriched and uncorrected, \rtwo enriched and SMC-corrected, 
and \rthree unenriched base RL.
Our goal is to compare the training trajectories of the three approaches, and 
analyze how the effect of enrichment we have established manifests itself in real training.

We train Qwen3-1.7B on a hard 500-problem subset of OpenMathReasoning where the 
pass@1 rate of the base policy is roughly 2\%, 
retain a held-out 300-problem set for evaluation, 
and perform GRPO with a group size of 16 for 500 updates.
Training results are averaged over three seeds.
Our enrichment comes from a parallel LLM that is \emph{hinted} with a partial solution, 
similar to~\citep{agrawal2026off}, 
asked to produce a derivation and an answer after seeing the hint.
We center our investigation on tracking policy performance across matched training 
steps.



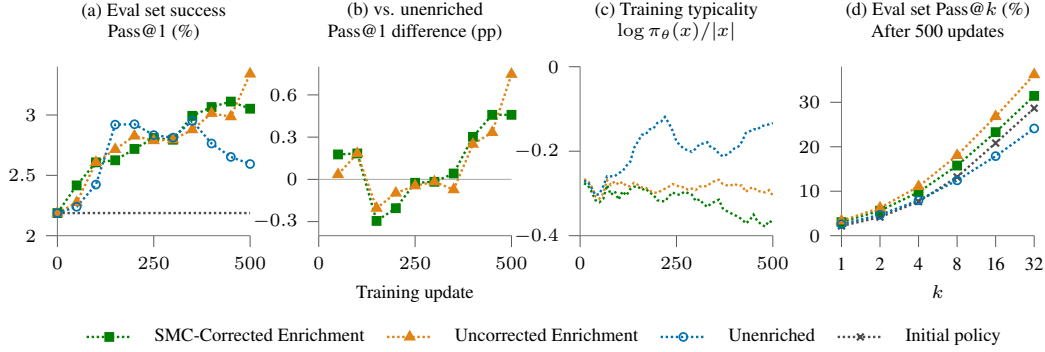
\begin{figure}[!t]
\centering
\begingroup
\pgfplotsset{
  math corrected/.style={paper corrected},
  math uncorrected/.style={paper uncorrected},
  math iid/.style={paper unenriched},
  math initial/.style={paper initial},
}
\begin{tikzpicture}
\begin{groupplot}[
  compact paper axis,
  /tikz/font=\scriptsize,
  title style={font=\scriptsize,align=center},
  group style={group size=4 by 1,horizontal sep=0.065\linewidth},
  width=0.1825\linewidth,height=0.16\linewidth,scale only axis,
  xmin=0,xmax=500,xtick={0,250,500},
  unbounded coords=jump,
  axis lines*=left,
  axis line style={black!55,line width=0.35pt},
  grid=none,
  tick label style={font=\scriptsize},
  /tikz/line width=0.9pt,
]
\nextgroupplot[
  title={(a) Eval set success\\Pass@1 (\%)},
  ymin=2,ymax=3.4,ytick={2,2.5,3},
  legend columns=4,
  legend style={at={(2.534,-0.48)},anchor=north,column sep=6pt,font=\scriptsize},
]
\addplot[math initial,no marks,forget plot] coordinates {(0,2.1875) (500,2.1875)};
\addplot[math corrected] table[x=step,y=pass1_corrected,col sep=comma]{figs/data/singleprob_training.csv};
\addlegendentry{SMC-Corrected Enrichment}
\addplot[math uncorrected] table[x=step,y=pass1_uncorrected,col sep=comma]{figs/data/singleprob_training.csv};
\addlegendentry{Uncorrected Enrichment}
\addplot[math iid] table[x=step,y=pass1_iid,col sep=comma]{figs/data/singleprob_training.csv};
\addlegendentry{Unenriched}
\addlegendimage{math initial}
\addlegendentry{Initial policy}
\nextgroupplot[
  title={(b) vs. unenriched\\Pass@1 difference (pp)},
  ymin=-0.4,ymax=0.8,ytick={-0.3,0,0.3,0.6},
]
\addplot[gray!65,thin,forget plot] coordinates {(0,0) (500,0)};
\addplot[math corrected] table[x=step,y=delta_corrected,col sep=comma]{figs/data/singleprob_training.csv};
\addplot[math uncorrected] table[x=step,y=delta_uncorrected,col sep=comma]{figs/data/singleprob_training.csv};
\nextgroupplot[
  title={(c) Training typicality\\$\log\pi_\theta(x)/|x|$},
  ymin=-0.4,ymax=0,ytick={-0.4,-0.2,0},
]
\addplot[math corrected,no marks] table[x=step,y=corrected,col sep=comma]{figs/data/singleprob_typicality.csv};
\addplot[math uncorrected,no marks] table[x=step,y=uncorrected,col sep=comma]{figs/data/singleprob_typicality.csv};
\addplot[math iid,no marks] table[x=step,y=iid,col sep=comma]{figs/data/singleprob_typicality.csv};
\nextgroupplot[
  title={(d) Eval set Pass@$k$ (\%)\\After 500 updates},
  xmode=log,log basis x=2,xmin=1,xmax=32,
  xtick={1,2,4,8,16,32},xticklabels={1,2,4,8,16,32},
  ymin=0,ymax=38,ytick={0,10,20,30},
]
\addplot[math initial] table[x=k,y=base,col sep=comma]{figs/data/singleprob_passk500.csv};
\addplot[math corrected] table[x=k,y=corrected,col sep=comma]{figs/data/singleprob_passk500.csv};
\addplot[math uncorrected] table[x=k,y=uncorrected,col sep=comma]{figs/data/singleprob_passk500.csv};
\addplot[math iid] table[x=k,y=iid,col sep=comma]{figs/data/singleprob_passk500.csv};
\end{groupplot}
\path (group c1r1.south) -- node[midway,below=15pt,font=\scriptsize] {Training update} (group c3r1.south);
\node[font=\scriptsize,anchor=north] at ([yshift=-15pt]group c4r1.south) {$k$};
\end{tikzpicture}
\endgroup
\caption{
Results of training Qwen3-1.7B on the described OpenMathReasoning subset, using 
\rone uncorrected enrichment, 
\rtwo SMC-corrected enrichment, and 
\rthree unenriched training.
Panel (a) maps pass@1 percentages on the eval set across training steps.
Panel (b) illustrates the advantage that uncorrected and SMC-corrected enrichment have 
compared to the unenriched baseline across training steps.
Panel (c) maps training typicality, the average length-normalized likelihood of the training samples 
according to each policy being trained, across training steps.
Panel (d) maps pass@$k$ across $k$ after all 500 training steps completed, illustrating that 
the two enriched arms achieve an increase compared to the base untrained policy while 
unenriched training results in a decrease.
}
\label{fig:singleprob}
\end{figure}

\Cref{fig:singleprob} illustrates the effects of enrichment.
Panel (a) illustrates pass@1 performance evolving as training proceeds, 
showing that both enriched training approaches do achieve a better 
terminal policy compared to unenriched training.
Panel (b) illustrates the difference between baseline unenriched training: 
at early training steps unenriched training outperforms 
both enriched trainers (roughly up to step 250), with 
the enriched trainers pulling ahead eventually by the terminal 500-th training step.

We explain this bimodal phenomenon via the analysis we have already established:
at earlier steps, 
unenriched training sees a small amount of rewarding samples from the base policy 
and consistently reinforces the behavior on this set.
This interpretation is supported by panel (c), which maps the 
typicality (length-normalized average log-probability)
of the training samples a policy $\policy$ sees, according to $\policy$ itself, across 
training timesteps.
Typicality grows until roughly 200 timesteps before plateauing, 
indicating that the
unenriched-trained policy 
finds its own training rollouts more likely compared to its starting point.
Panel (d) then shows that pass@k rate has 
\emph{decreased} compared to the base policy for unenriched training, 
illustrating classical RL collapse: 
the unenriched trainer's sparse successes are concentrated on a few problems, 
and RL reinforces this narrow band of successes such that 
learning seems to proceed in early training steps (panel (a)) but eventually 
collapses towards memorizing a set of known solutions (the pass@k decrease in panel (d)).

On the other hand, the collapse does \emph{not} occur for the two enriched trainers: 
in panel (c), both uncorrected and SMC-corrected show a mild decrease in typicality.
This decrease comes from the policy's own entropy increasing
hinted enrichment 
feeds it a diverse range of successful samples that the base policy cannot generate: 
the policy is learning new successes.
In particular, panel (d) indicates that
even the SMC-corrected arm has successfully increased pass@k, 
indicating that exposing the model to enriched successes has helped \emph{prevent} collapse 
(even though the corrected weights are small).
We do note that the difference between SMC-corrected and unenriched 
training was statistically insignificant at a 
95\% sample bootstrap interval~\citep{taylor} of $[-0.21\text{pp}, +1.12\text{pp}]$ 
(a one-sided test of $($SMC $>$ unenriched$)$ yields $p=0.09$).
On the other hand, the pass@$k$ increase between SMC and unenriched was statistically significant, 
indicating that although SMC maybe improving slower due to the corrected gradients, 
it is definitely avoiding the mode of collapse displayed by unenriched training.

\textbf{Conclusion.}
Enrichment can be helpful for real training, especially in very sparse domains, 
where enrichment provides access to \emph{new} successes the base policy would have failed to 
generate otherwise.
The gradient distortion of uncorrected consumption is not necessarily harmful in this regime, 
especially as the distortion points towards accelerating learning 
on successes not in the base policy.
Corrected consumption also benefits from having access to the new successes, 
establishing 
a concrete example domain and analysis of when enrichment can help real-world RL.

\section{Discussion and Related Work}
\label{sec:related}


\textbf{Strategies for enrichment.}
\citet{zhang2026bread,huang2025blending,zhang2025adhint} 
use expert prefixes or hints to make successful continuations more accessible during training, 
similar to the hinted approach we used in \Cref{sec:results}.
PaTR~\citep{li2026process} instead uses process feedback to branch from promising 
intermediate states and stop unproductive paths.
Resampling methods, 
such as sequential resampling, similarly reallocate computation among partial trajectories,
with auxiliary particle filters~\citep{pitt1999filtering} 
and weighted ensemble sampling~\citep{aristoff2018analysis} 
providing precedents for selection weights.
Twisted SMC for language models~\citep{zhao2024probabilistic} 
can also be understood, in general, as an enrichment strategy (where the enrichment is provided 
by the twist) 
to reliably produce a set of samples that satisfy a certain property
(e.g., having a reward).
We apply the idea of twisted SMC to estimate
corrected gradients for training a policy from enriched rollouts.

\textbf{Learning from selected successes.}
STaR~\citep{zelikman2022star} and ReST$^{\mathrm{EM}}$~\citep{singh2023beyond} 
perform fine-tuning on successful responses that the model generates instead of performing 
reinforcement learning.
Similarly, 
ReST-MCTS*~\citep{zhang2024rest} uses process-reward-guided tree search to collect training traces.
These methods explicitly learn from selected examples, 
rather than requiring an unbiased estimate of the base-policy gradient.
We observe that uncorrected consumption of enrichment is equivalent 
(under scale normalization)
to fine-tuning on the 
successful samples of an enriched policy; a derivation is provided in \Cref{app:enrichment-sft}.

\textbf{Importance correction.}
\citet{pmlr-v80-espeholt18a} uses truncated importance weights to 
handle actor--learner policy mismatch, 
illustrating the longstanding trade-off between correction and variance.
Off-Context GRPO~\citep{agrawal2026off} addresses 
the mismatch between guided generation and unguided training through token-level importance ratios.
Active importance sampling for policy gradients~\citep{papini2024policy} 
explicitly optimizes the sampling policy to reduce gradient variance.
In LLM RL, importance correction has recently been revisited: \citet{yao2026group} show
that group-relative REINFORCE is already off-policy, \citet{luo2025llm}
and \citet{zhang2026rethinking} study sequence versus token-level ratios,
and \citet{huang2026stable} control off-policy variance via effective sample size.
Our focus is how enrichment and its correction change learning, 
and in \Cref{sec:results} we 
connect our analysis to a real Qwen3-1.7B training run to explain the 
effects of enrichment across training.

\section{Conclusion}
\label{sec:conclusion}

This paper studies enrichment itself as a component of RL. 
We mathematically establish enrichment as an implicit reweighting of rewards, 
clarify the effect that this reweighting can have on training, 
introduce SMC-correction to variance in weight correction, and 
analyze the effects of enrichment on a real training run.
An interesting line of future work would be to develop an adaptive RL 
mechanism that looks at online policy behavior to 
choose the training strategy,
based on the results in this paper.

\subsection*{AI use statement}

In this work, we used generative AI tools to help formulate the 
theorems in proofs in \Cref{sec:2analysis,sec:smc} and the Appendices,
design, implement and analyze the experiments in \Cref{sec:results},
investigate related work in \Cref{sec:related},
and improve figures and writing.
We have reviewed all AI-assisted work.
We take responsibility for the final content of this work,
including text, claims or artifacts produced with the aid of generative AI.

\subsection*{Ethics statement}

This paper presents work whose goal is to advance the field
of Machine Learning. 
We do not believe that our work raises questions related to the Code of Ethics.

\subsection*{Reproducibility statement}

Our work primarily concerns a mathematical analysis of the effects of enrichment.
The code used to run the experiment in \Cref{sec:results} and our implementation of the 
SMC corrector cannot be released at the time of review due to corporate licensing issues, 
but we plan to open-source the code upon publication.

\bibliography{iclr2027_conference}
\bibliographystyle{iclr2027_conference}

\newpage
\appendix
\section{GRPO and Gradient Distortion}
\label{app:grpo-distortion}

In \Cref{sec:2analysis} we discussed how 
enrichment implicitly reweights rewards and distorts the gradient in a base 
REINFORCE setting. 
Here we extend the analysis to GRPO.

\subsection{Reweighted Reward in GRPO}
\label{app:grpo-reweighted-reward}

Fix one problem and parameters $\param$, and draw $N\geq2$ independent responses $\sample_1,\ldots,\sample_N\sim\qpolicy$.
We consider the response-level GRPO reward update at these rollout parameters, without clipping, length normalization, or KL regularization.
Assume $\bpolicy(\sample)>0$ wherever $\qpolicy(\sample)>0$, and that the expectations below are finite.
Define the group reward mean $\overline{\reward}$, sample standard deviation $s_{\reward}$, and advantages $\widehat A_i$ as in GRPO~\citep{shao2024deepseekmath}:
{\small
\begin{equation}
  \overline{\reward}=\frac1N\sum_{j=1}^{N}\reward(\sample_j),
  \qquad
  s_{\reward}^{2}=\frac1{N-1}\sum_{j=1}^{N}\bigl(\reward(\sample_j)-\overline{\reward}\bigr)^2,
  \qquad
  \widehat A_i=\frac{\reward(\sample_i)-\overline{\reward}}{s_{\reward}+\epsilon},
  \label{eq:grpo-distortion-advantages}
\end{equation}
}
where $\epsilon>0$.
Treating the advantages as fixed coefficients when differentiating, the uncorrected enriched update is
{\small
\begin{equation}
  \widehat u_{\mathrm{GRPO}}
  =\frac1N\sum_{i=1}^{N}\widehat A_i\nabla_\param\log\bpolicy(\sample_i).
  \label{eq:grpo-distortion-update}
\end{equation}
}
Each group position has the same expected contribution, so averaging over the full sampled group and then conditioning on $\sample_1$ gives
{\small
\begin{equation}
\begin{aligned}
  \e_{\sample_1,\ldots,\sample_N\overset{\mathrm{i.i.d.}}{\sim}\qpolicy}[\widehat u_{\mathrm{GRPO}}]
  &=\e_{\sample_1,\ldots,\sample_N\overset{\mathrm{i.i.d.}}{\sim}\qpolicy}\!\left[\widehat A_1\nabla_\param\log\bpolicy(\sample_1)\right]\\
  &=\sum_{\sample}\qpolicy(\sample)\,
    \e\!\left[\widehat A_1\mid\sample_1=\sample\right]
    \nabla_\param\log\bpolicy(\sample)\\
  &=\e_{\sample\sim\bpolicy}\!\Bigl[
    \underbrace{\frac{\qpolicy(\sample)}{\bpolicy(\sample)}
      \e\!\left[\widehat A_1\mid\sample_1=\sample\right]}_{\displaystyle{\rreward}^{\mathrm{GRPO}}(\sample)}
    \nabla_\param\log\bpolicy(\sample)
    \Bigr].
  \label{eq:grpo-distortion-reweighting}
\end{aligned}
\end{equation}
}
The conditional expectation averages over the remaining $N-1$ independent responses from $\qpolicy$.
The outer sum averages over all possible values of $\sample_1$.
Thus, the effective reward ${\rreward}^{\mathrm{GRPO}}$ 
can be understood as 
reweighting the expected group-relative advantage, as opposed to the raw reward $\reward$
(similar to how GRPO replaces the raw reward with the group-relative advantage).
As in \Cref{sec:reward-reweighting}, this reweighting factor depends on the current network 
parameters $\param$ and is not fixed throughout training.
The decomposition into scale, rotation, and variance, and their effects on training
apply identically, but are realized under 
different conditions.
Let $h = \e[\widehat{u}_{\mathrm{GRPO}}]$ and assume $g \neq 0$.

\paragraph{Scale.}
GRPO induces pure scale bias when the 
reweighted expected advantages satisfy:
{\small
\begin{equation}
  {\rreward}^{\mathrm{GRPO}}(\sample)
  =\alpha\reward(\sample)+c
  \qquad\text{for every }\sample,
  \label{eq:grpo-distortion-pure-scale}
\end{equation}
}
Here $c$ is independent of the response $\sample$ and $\alpha=\langle h,g\rangle/\|g\|^2$.
Then $h=\alpha g$, since $\e_{\sample\sim\bpolicy}[\nabla_\param\log\bpolicy(\sample)]=0$ eliminates the constant term.

\paragraph{Rotation.}
A sufficient condition for pure rotation is to transfer equal amounts of expected contribution between two responses whose score gradients differ but have the same projection onto $g$.
Specifically, suppose that, for responses $\sample_a,\sample_b$, a nonzero $\delta$, and a response-independent $c$,
{\small
\begin{equation}
  {\rreward}^{\mathrm{GRPO}}(\sample)
  =\reward(\sample)+c
    +\delta\left(
      \frac{\mathbf{1}\{\sample=\sample_a\}}{\bpolicy(\sample_a)}
      -\frac{\mathbf{1}\{\sample=\sample_b\}}{\bpolicy(\sample_b)}
    \right).
  \label{eq:grpo-distortion-rotation-transfer}
\end{equation}
}
Then the mean update is
{\small
\begin{equation}
  h=g+\delta\nabla_\param\log
    \frac{\bpolicy(\sample_a)}{\bpolicy(\sample_b)}.
  \label{eq:grpo-distortion-rotation-update}
\end{equation}
}
If $\langle g,\nabla_\param\log\bpolicy(\sample_a)\rangle=\langle g,\nabla_\param\log\bpolicy(\sample_b)\rangle$ but the two score gradients differ, the added component is nonzero and perpendicular to $g$.
Thus $\alpha=1$ and $r\neq0$: there is rotation error but no scale error in the decomposition of \Cref{sec:gradient-decomposition}.

\paragraph{Variance.}
Consider binary rewards with success probability $p=\p_{\sample\sim\qpolicy}(\reward(\sample)=1)\in(0,1)$.
The probability that a group contains both successes and failures is $m=1-(1-p)^N-p^N$; every other group has identical rewards and hence zero advantages and zero update.
Conditioning on whether the group is mixed gives
{\small
\begin{equation}
  \e\!\left[\norm{\widehat u_{\mathrm{GRPO}}-h}^{2}\right]
  \geq\frac{1-m}{m}\norm{h}^{2}.
  \label{eq:grpo-distortion-rare-success-variance}
\end{equation}
}
For example, with $N=128$ and $p=10^{-4}$, approximately $98.7\%$ of groups produce no update, and the variance is at least $77.6\norm{h}^{2}$.
For a nonzero mean $h$, this is high variance relative to the squared mean update: the expected learning signal is carried by rare mixed groups, rather than appearing consistently across updates.

\subsection{GRPO Distorts Training Relative to Pure Reinforce}
\label{app:grpo-unenriched-distortion}

In addition to the reweighted reward derivation in \Cref{app:grpo-reweighted-reward}, 
here we note that 
GRPO updates distort training relative to pure REINFORCE training, 
\emph{even without} enrichment.
We derive the distortion 
by observing that GRPO induces a \emph{per-problem} scale factor, 
which can further induce rotation distortion when aggregating gradients over multiple problems.

Set $\qpolicy=\bpolicy$ and assume binary rewards, with gradients computed by
taking the sum over token log-probability gradients, 
multiplied by the corresponding group-relative advantage, and 
averaging over responses.
For one problem, let $p_\param=\e_{\sample\sim\bpolicy}[\reward(\sample)]\in(0,1)$ 
be its success probability, 
and write $\widehat g_{\mathrm{GRPO}}$ for the GRPO update from $N$ independent base-policy responses.
The true reference gradient is $g=\e_{\sample\sim\bpolicy}[\reward(\sample)\nabla_\param\log\bpolicy(\sample)]$.

\paragraph{Per-problem scale.}
Using $\e_{\sample\sim\bpolicy}[\nabla_\param\log\bpolicy(\sample)]=0$, the conditional score means are
{\small
\begin{equation}
\begin{aligned}
  \e_{\sample\sim\bpolicy}\!\left[
    \nabla_\param\log\bpolicy(\sample)\mid\reward(\sample)=1
  \right]&=\frac{g}{p_\param},\\
  \e_{\sample\sim\bpolicy}\!\left[
    \nabla_\param\log\bpolicy(\sample)\mid\reward(\sample)=0
  \right]&=-\frac{g}{1-p_\param}.
\end{aligned}
\label{eq:grpo-iid-conditional-scores}
\end{equation}
}
Let $K=\sum_{i=1}^{N}\reward(\sample_i)\sim\mathrm{Binomial}(N,p_\param)$ be the number of successes in the group.
Conditional on $K=k$, the reward mean is $k/N$ and the sample standard deviation is $s_k=\sqrt{k(N-k)/(N(N-1))}$.
Substituting the positive and negative advantages gives
{\small
\begin{equation}
\begin{aligned}
  \e[\widehat g_{\mathrm{GRPO}}\mid K=k]
  &=\frac{k}{N}\frac{1-k/N}{s_k+\epsilon}\frac{g}{p_\param}
    +\frac{N-k}{N}\frac{-k/N}{s_k+\epsilon}
      \frac{-g}{1-p_\param}\\
  &=\frac{k(N-k)}{N^2(s_k+\epsilon)p_\param(1-p_\param)}\,g.
\end{aligned}
\label{eq:grpo-iid-count-conditioned-update}
\end{equation}
}
The all-failure and all-success groups ($k=0,N$) contribute zero.
Averaging over all groups therefore yields
{\small
\begin{equation}
\begin{aligned}
  \e_{\sample_1,\ldots,\sample_N\overset{\mathrm{i.i.d.}}{\sim}\bpolicy}
    [\widehat g_{\mathrm{GRPO}}]
    &=\alpha_N(p_\param)\,g,\\
  \alpha_N(p_\param)
    &=\frac{1}{p_\param(1-p_\param)}
      \sum_{k=1}^{N-1}\binom Nk p_\param^k(1-p_\param)^{N-k}
      \frac{k(N-k)}{N^2(s_k+\epsilon)}.
\end{aligned}
\label{eq:grpo-iid-scale-factor}
\end{equation}
}
The factor $\alpha_N(p_\param)$ is positive but need not equal one: unenriched GRPO can introduce scale bias even on one problem, without rotating that problem's expected update.

\paragraph{Rotation across problems.}
Now consider two equally weighted problems at the same parameters $\param$, with success probabilities $p_{1,\param},p_{2,\param}$ and true gradients $g_1,g_2$.
The aggregate true gradient $g_{\mathrm{all}}$ and expected GRPO update $h_{\mathrm{all}}$ are
{\small
\begin{equation}
  g_{\mathrm{all}}=\frac{g_1+g_2}{2},
  \qquad
  h_{\mathrm{all}}=\frac{
    \alpha_N(p_{1,\param})g_1+\alpha_N(p_{2,\param})g_2}{2}.
  \label{eq:grpo-iid-across-problem-rotation}
\end{equation}
}
If $g_1,g_2$ are linearly independent and their scale factors differ, no single scalar maps $g_{\mathrm{all}}$ to $h_{\mathrm{all}}$, so the aggregate update has rotation error.
For example, with $N=16$ and $\epsilon=10^{-6}$, success probabilities $0.05$ and $0.5$ give scale factors approximately $3.16$ and $1.87$, respectively.
Thus, per-problem normalization can change the relative contributions of problems 
and rotate their combined gradient, even without enrichment or importance correction.

\section{SMC Correction}
\label{app:smc-correction}

In this section we provide extra information and proofs about our SMC-corrector 
in \Cref{sec:smc}.

\subsection{Dual Weights for Enforcing Enrichment}
\label{app:smc-guidance}

We observe that the mechanism presented in \Cref{sec:smc-sequential} reduces variance, but 
can counterbalance the effects of guidance.
For example, if $\qpolicy$ is a proper enrichment, 
then the correction weight $W_i$ will be low for rewarding samples preferred by $\qpolicy$ and 
high for non-rewarding samples preferred by $\bpolicy$: counteracting the enrichment 
at a resampling step.
To alleviate this tension, we extend the presented mechanism by attaching 
an \emph{additional enrichment weight} $\bweight_i > 0$ to each sample $x_i$.
The enrichment weight $\bweight_i$ tracks, intuitively, 
the expected future reward for $x_i$:
for example, using a process reward model (PRM) score.
We observe that $\bweight_i$ need not be tracked iteratively like $\cweight_i$; 
it suffices that $\bweight_i$ is computable from the partial sample $x_i$.
In this section, we assume that $\bweight_i$ is a suitable proxy for expected future reward; 
the SMC analogue for $\bweight_i$ is the resample potential~\cite{zhao2024probabilistic}.

We trigger resampling by tracking the variance of the product weight $\cweight_i \bweight_i$, 
where the weighted resample also uses the product weights for resampling:
intuitively, multiplying the enrichment weight $\bweight_i$
restores the disadvantage that a sample $x_i$ may have accumulated in 
$\cweight_i$ due to a proper enrichment.
At a resample, a child sample $\sample_c$ originating from $\sample_i$ then inherits:
{\small
\begin{equation}
  \cweight_c=\frac{\sum_j \cweight_j \bweight_j}{N} \cdot  \frac{1}{\bweight_i}
  \qquad
  \bweight_c=\bweight_i.
  \label{eq:smc-reset}
\end{equation}
}
Here the average is over the product weights $\cweight_j\bweight_j$, rather than the correction weights alone.
The enrichment weight $\bweight_i$ is preserved because children inherit the same prefix.
The intuition is that the enrichment weight
increases the importance of the sample $x_i$ by a factor of $\bweight_i$, which results in more clones; 
thus the children of $x_i$ after resampling must have a lower 
correction weight $\cweight$ to account for the additional samples.
Once generation is complete, we discard the enrichment weights $\bweight_i$ and use only 
the correction weights $\cweight_i$ to compute the gradient: 
after all $N$ samples are generated, 
$\hat{g}_{\mathrm{SMC}}$, the dual-weight SMC gradient estimator, is defined as:
{\small
\begin{equation}
\hat{g}_{\mathrm{SMC}}
  =\frac1N\sum_{i=1}^{N}
    \cweight_i\reward(\sample_i)\nabla_\param\log\bpolicy(\sample_i)
\label{eq:smc-gradient}
\end{equation}
}

\subsection{Mean Preservation and Conditions for Additive Variance Accumulation}
\label{app:smc-additive-variance}

We prove that the SMC-corrected REINFORCE estimator preserves the true gradient in expectation, then derive conditions for an additive-in-length variance bound.
Mean preservation does not require these additional variance-control conditions.
In particular, an additive decomposition of the error does not by itself prevent its terms from growing exponentially with response length.

We first isolate the effect of resampling on correction weights, 
then derive a finite-population bound for the gradient estimator in \Cref{eq:smc-gradient}.
We observe that the steps must be separate, because standard weight correction 
only does not account for the fact that two separate samples inside an SMC-sample batch 
can share the same prefix ancestry.
We analyze the dual-weight rule in \Cref{sec:smc} with ESS-triggered systematic resampling, 
using conditional-variance accounting for particle estimators~\citep{chopin}.

\paragraph{Assumptions.}
Let us fix one problem and parameters $\param$ during rollout, 
with a finite vocabulary and response length bounded by $l$.
Assume finished responses are padded by a token of probability $1$ under both policies, with correction ratio $1$ and zero score gradient.
All $N$ particles start at the empty prefix with correction weight $\cweight_i=1$.
The base and enriched policies have common positive support, 
and all importance ratios are exact, without clipping or self-normalization.
We use three bounds, each controlling a different part of the estimator.
\begin{subequations}
\label{eq:smc-variance-assumptions}
First, for a next token $a$ following prefix $s$, define the token-level correction ratio $\omega_\param(a\mid s)$ and assume that its variance under the enriched policy is at most $v$:
{\small
\begin{equation}
  \omega_\param(a\mid s)
    :=\frac{\bpolicy(a\mid s)}{\qpolicy(a\mid s)},
  \qquad
  \operatorname{Var}_{a\sim\qpolicy(\cdot\mid s)}
    [\omega_\param(a\mid s)]\leq v.
\end{equation}
}
The ratio has conditional mean $1$ by normalization of $\bpolicy$; the assumption bounds the variability introduced by one token, not the variance of a full response's correction weight.
The same constant $v$ must apply at every reachable prefix.

Second, assume that the enrichment weight $\bweight(s)$ is bounded above and bounded away from zero, and define $\rho$ as the ratio of these bounds:
{\small
\begin{equation}
  0<b_{\min}\leq\bweight(s)\leq b_{\max}<\infty,
  \qquad
  \rho:=\frac{b_{\max}}{b_{\min}}.
\end{equation}
}
The positive lower bound makes division by $\bweight(s)$ in the resampling reset well-defined, while $\rho$ controls how much enrichment weights can differ across prefixes.

Third, assume there is a finite constant $H$ bounding the norm of the reward-weighted score gradient for every complete response $\sample$:
{\small
\begin{equation}
  \norm{\reward(\sample)\nabla_\param\log\bpolicy(\sample)}\leq H.
\end{equation}
}
This bounds the magnitude of a response's REINFORCE contribution before multiplying by its correction weight $\cweight$.
The bound $H$ may depend on the response length $l$: the score is a sum of token-level gradients.
We retain this dependence when interpreting the horizon-dependent variance bound below.
\end{subequations}
These are sufficient conditions, not properties established for arbitrary enriched LLM policies.
The full-gradient variance bound will additionally require the particle-order condition in \Cref{eq:smc-order-regularity}.

\paragraph{ESS-triggered weighted resampling.}
Following the token-level presentation in \Cref{sec:smc}, we check ESS after each token except the last and resample only when
{\small
\begin{equation}
  \operatorname{ESS}(\cweight\bweight)
  :=\frac{(\sum_i\cweight_i\bweight_i)^2}
          {\sum_i(\cweight_i\bweight_i)^2}
  <\tau N
\label{eq:smc-ess-trigger}
\end{equation}
}
for a fixed trigger threshold $0<\tau\leq1$.
When resampling fires, we perform weighted resampling to select $N$ parents, using normalized weights $p_i=\cweight_i\bweight_i/\sum_j\cweight_j\bweight_j$.
If $C_i$ counts the children of parent $i$, the resampling satisfies~\citep{smcresampling}
{\small
\begin{equation}
  \sum_i C_i=N,
  \qquad \e[C_i\mid\mathcal F]=Np_i,
  \qquad C_i\in\{\lfloor Np_i\rfloor,\lceil Np_i\rceil\},
\label{eq:smc-systematic-counts}
\end{equation}
}
Here $\mathcal F$ denotes the sampling history before this resampling operation, including the current population and its order.
Resampling uses fresh randomness conditional on this history; new token draws are conditionally independent across particles.
The offspring counts need not be independent; the variance calculation below accounts for their dependence.
Each child receives the reset in \Cref{eq:smc-reset}.

After a skipped check, ESS is at least $\tau N$; after resampling, all child products $\cweight_i\bweight_i$ are equal, so ESS is $N$.
Writing $\overline{\cweight}=N^{-1}\sum_i\cweight_i$, the bound on $\bweight$ therefore gives, before every generation operation,
{\small
\begin{equation}
  \sum_i\cweight_i^2
  \leq\frac{\sum_i(\cweight_i\bweight_i)^2}{b_{\min}^2}
  \leq\frac{(\sum_i\cweight_i\bweight_i)^2}{b_{\min}^2\tau N}
  \leq\frac{\rho^2}{\tau}\,N\overline{\cweight}^{\,2}.
\label{eq:smc-ess-weight-control}
\end{equation}
}
Initialization also satisfies this inequality.
\Cref{eq:smc-ess-weight-control} is important, as it essentially bounds 
the variance on the weights after resampling, which we will use below.


\paragraph{Averaging before multiplying.}
First set $\bweight=1$ and retain the assumption that each token's correction ratio has conditional variance at most $v$.
Each ratio has conditional mean $1$ because $\sum_a\qpolicy(a\mid s)\omega_\param(a\mid s)=1$, so its expected square is at most $1+v$.
Without resampling, a response's terminal weight $\cweight_i$ is the product of its $l$ ratios, including the ratio-$1$ padding tokens for responses that finish early.
Applying the bound successively at each token gives $\e[\cweight_i^2]\leq(1+v)^l$.
The expected squared weight $\e[\cweight_i^2]$ is called its \emph{second moment}; since $\e[\cweight_i]=1$, its variance is $\e[\cweight_i^2]-1$.
For $N$ independent responses, the variance of the average terminal weight is therefore bounded by
{\small
\begin{equation}
  \operatorname{Var}\!\left[
    \frac1N\sum_{i=1}^{N}\prod_{k=1}^{l}\omega_{\param,k,i}
  \right]
  \leq\frac{(1+v)^l-1}{N},
  \qquad
  \omega_{\param,k,i}
  :=\omega_\param(t_{k,i}\mid t_{1:k-1,i}).
\label{eq:smc-independent-weight-variance}
\end{equation}
}
With ESS-triggered resampling, the weights need not be equal before each token, but \Cref{eq:smc-ess-weight-control} bounds their concentration.
Let $\overline{\cweight}_k$ be the mean weight after token $k$ and its optional resampling operation.
When $\bweight=1$, resampling preserves the mean weight exactly, irrespective of which parents are selected.
Conditioning on the population $\mathcal F$ before token $k$, independence of the new token draws gives
{\small
\begin{equation}
  \e[\overline{\cweight}_k^2\mid\mathcal F]
  \leq\overline{\cweight}_{k-1}^{\,2}
    +\frac{v}{N^2}\sum_i\cweight_i^2
  \leq\left(1+\frac{v}{\tau N}\right)
    \overline{\cweight}_{k-1}^{\,2}.
\label{eq:smc-mean-weight-recursion}
\end{equation}
}
Applying this inequality successively over the $l$ token positions,
starting from \(\overline W_0=1\), bounds the expected squared terminal weight. 
Since its mean is 1, subtracting 1 gives the variance bound:
{\small
\begin{equation}
  \operatorname{Var}[\overline{\cweight}_l]
  \leq\left(1+\frac{v}{\tau N}\right)^l-1
  \leq e^{lv/(\tau N)}-1.
\label{eq:smc-weight-additive-regime}
\end{equation}
}
In the SMC bound, each new token's variance contribution is divided by at least $\tau N$ \emph{before} it compounds with earlier randomness 
(the $v/\tau N$ term in the middle equation), 
rather than averaging $N$ already-compounded weights at the end.
The upper bound is $lv/(\tau N)+O((lv/(\tau N))^2)$ when $lv/(\tau N)$ is small: its leading term accumulates additively over tokens.
For example, when $lv\leq\tau N$, the variance is at most $(e-1)lv/(\tau N)$.
At fixed $N$, the bound remains exponential for sufficiently large $l$; the linear regime requires $lv/(\tau N)$ to remain bounded.
Thus resampling reduces the rate of compounding in this bound, rather than removing exponential growth at every horizon.

\paragraph{Accounting for the full gradient.}
To convert the variance over correction weights into a full variance result over gradients, 
define the expected final gradient contribution from prefix $s$ as
{\small
\begin{equation}
  m_\param(s)
  :=\e_{\sample\sim\bpolicy}\!\left[
    \reward(\sample)\nabla_\param\log\bpolicy(\sample)
    \mid\sample_{1:|s|}=s
  \right].
\label{eq:smc-prefix-gradient-expectation}
\end{equation}
}
\Cref{eq:smc-prefix-gradient-expectation} satisfies $m_\param(\emptyset)=g$ at the empty prefix, $m_\param(\sample)=\reward(\sample)\nabla_\param\log\bpolicy(\sample)$ for a complete response, and $\norm{m_\param(s)}\leq H$.
For a current population of prefixes $s_i$, define its average correction weight $\overline{\cweight}$, average selection weight $\overline{\cweight\bweight}$, and weighted expected gradient $M$ by
{\small
\begin{equation}
  \overline{\cweight}=\frac1N\sum_i\cweight_i,
  \qquad
  \overline{\cweight\bweight}=\frac1N\sum_i\cweight_i\bweight_i,
  \qquad
  M=\frac1N\sum_i\cweight_i m_\param(s_i).
\label{eq:smc-population-prediction}
\end{equation}
}
We condition on the population $\mathcal F$ immediately before each operation, holding all prefixes $s_i$ and weights $\cweight_i,\bweight_i$ fixed.
The remaining randomness is in the next-token draws for generation, or in the offspring counts $C_i$ for resampling.
Averaging over that randomness leaves $M$ unchanged:
{\small
\begin{equation}
\begin{aligned}
  \sum_a\qpolicy(a\mid s)\omega_\param(a\mid s)m_\param(sa)
    &=m_\param(s),\\
  \e[M^{\mathrm{res}}\mid\mathcal F]
    =\frac{\overline{\cweight\bweight}}{N}
      \sum_i\frac{\e[C_i\mid\mathcal F]}{\bweight_i}m_\param(s_i)
    &=M.
\end{aligned}
\label{eq:smc-prediction-preservation}
\end{equation}
}
Here $M^{\mathrm{res}}$ is the weighted expected gradient after resampling.
The second equality uses expected offspring counts, not independent parent draws.
A skipped resampling check leaves $M$ unchanged, and the ESS decision is determined by the current population, so the adaptive schedule also preserves the conditional mean.

To accumulate the errors over a full rollout, number the $l$ generation operations and $l-1$ resampling checks by $j=1,\ldots,2l-1$ in the order they occur.
A skipped check still receives an index, but leaves the population unchanged.
Write $M_j$ for $M$ evaluated on the population after operation $j$, and let $\mathcal F_j$ denote the sampling history up to that point.
This history records the tokens drawn and parents selected so far, and therefore determines all current prefixes and weights.
Conditioning on $\mathcal F_{j-1}$ means holding this history fixed and averaging only over the new random choices in operation $j$.
The preceding mean-preservation identities then give $\e[M_j\mid\mathcal F_{j-1}]=M_{j-1}$: the change $M_j-M_{j-1}$ has conditional mean zero.
Initially $M_0=g$, and after the final generation operation $M_{2l-1}=\hat g_{\mathrm{SMC}}$.
Taking expectations successively over these operations therefore gives
{\small
\begin{equation*}
  \e[\hat g_{\mathrm{SMC}}]
  =\e[M_{2l-1}]
  =\cdots
  =\e[M_0]
  =g.
\end{equation*}
}
Thus the SMC-corrected REINFORCE estimator is unbiased for the true gradient $g$, 
as stated in \Cref{sec:smc}.
This conclusion follows from mean preservation alone and does not require the 
variance bounds derived below.

This property also makes the cross terms between different operations vanish in expectation.
For $j<k$, fixing the history $\mathcal F_{k-1}$ fixes the earlier change $M_j-M_{j-1}$, while the later change $M_k-M_{k-1}$ still has conditional mean zero.
Their expected inner product is therefore zero: $\e[\langle M_j-M_{j-1},M_k-M_{k-1}\rangle]=0$.
The final error $\hat g_{\mathrm{SMC}}-g$ is the sum of these changes, so expanding its squared norm and taking expectations gives
{\small
\begin{equation}
  \e\!\left[\norm{\hat g_{\mathrm{SMC}}-g}^2\right]
  =\sum_{j=1}^{2l-1}\e\!\left[\norm{M_j-M_{j-1}}^2\right].
\label{eq:smc-additive-gradient-accounting}
\end{equation}
}
This identity includes resampling noise and the dependence caused by shared ancestors.
An additive decomposition alone does not imply linear growth: its individual terms must also remain controlled.

\paragraph{Bounding systematic resampling noise.}
The ESS bound controls the weights, but not the correlations between offspring counts.
By \Cref{eq:smc-systematic-counts}, each $C_i$ takes at most two adjacent integer values, so $\operatorname{Var}(C_i\mid\mathcal F)\leq1/4$.
For any real coefficients $z_1,\ldots,z_N$, Cauchy--Schwarz therefore gives
{\small
\begin{equation}
\begin{aligned}
  \operatorname{Var}\!\left[\sum_i z_i C_i\mid\mathcal F\right]
  &=\e\!\left[\left(\sum_i z_i(C_i-Np_i)\right)^2\mid\mathcal F\right]\\
  &\leq\left(\sum_i z_i^2\right)
    \sum_i\operatorname{Var}(C_i\mid\mathcal F)
  \leq\frac N4\sum_i z_i^2.
\end{aligned}
\label{eq:smc-count-covariance-bound}
\end{equation}
}
This bound includes correlations between offspring counts and does not assume independent parent selections.
It need not provide a variance reduction as $N$ grows: systematic resampling can correlate many offspring counts through its shared random offset~\citep{smcresampling}.
For example, with even $N$, alternating values $Np_i=3/2,1/2$ and coefficients $z_i=1,-1$ give variance $N^2/4$, attaining \Cref{eq:smc-count-covariance-bound}.
We therefore state an additional condition on the ordered population, rather than infer a $1/N$ factor from ESS alone.

\paragraph{Additional assumption for additive variance: controlled variation in particle order.}
At each triggered resample, index the parents $s_1,\ldots,s_N$ in the order used by systematic resampling.
Assume that a constant $K_{\mathrm{ord}}>0$, independent of $N$, $l$, and the resampling time, bounds the total change between neighboring parents as follows:
{\small
\begin{equation}
\begin{aligned}
  \sum_{i=1}^{N-1}
    \left\|\frac{m_\param(s_{i+1})}{\bweight_{i+1}}
            -\frac{m_\param(s_i)}{\bweight_i}\right\|
    &\leq 2\sqrt{K_{\mathrm{ord}}N}\,\frac{H}{b_{\min}},\\
  \sum_{i=1}^{N-1}
    \left|\frac1{\bweight_{i+1}}-\frac1{\bweight_i}\right|
    &\leq\sqrt{K_{\mathrm{ord}}N}
      \left(\frac1{b_{\min}}-\frac1{b_{\max}}\right).
\end{aligned}
\label{eq:smc-order-regularity}
\end{equation}
}
The first inequality controls variation in the resampled gradient contributions; the second controls variation in the resampled total correction weight.
They allow total variation to grow as $\sqrt N$, but exclude repeated fixed-amplitude alternation such as the counterexample above.
For example, in one dimension, if both coefficient sequences are monotone, their total variations are at most $2H/b_{\min}$ and $1/b_{\min}-1/b_{\max}$, respectively, so $K_{\mathrm{ord}}=1$ suffices.
The reciprocal-weight condition is automatic when $\bweight=1$.

To see their effect, let $D_k$ be the discrepancy between the total offspring of the first $k$ parents and its conditional expectation:
{\small
\begin{equation}
  D_k:=\sum_{i=1}^{k}(C_i-Np_i),
  \qquad D_N=0,
  \qquad \e[D_k\mid\mathcal F]=0,
  \qquad \e[D_k^2\mid\mathcal F]\leq\frac14.
\label{eq:smc-cumulative-count-discrepancy}
\end{equation}
}
The last bound follows because systematic resampling rounds each cumulative offspring count to one of the two integers surrounding $N\sum_{i=1}^{k}p_i$.
For any fixed vector coefficients $z_i$, summation by parts gives $\sum_i(C_i-Np_i)z_i=\sum_{k=1}^{N-1}D_k(z_k-z_{k+1})$.
The triangle inequality for the conditional root-mean-square norm therefore yields
{\small
\begin{equation}
  \e\!\left[\left\|\sum_i(C_i-Np_i)z_i\right\|^2\mid\mathcal F\right]
  \leq\frac14\left(\sum_{i=1}^{N-1}\|z_{i+1}-z_i\|\right)^2.
\label{eq:smc-ordered-count-variance}
\end{equation}
}
This bound retains the dependence between offspring counts; no independence assumption is introduced.

\paragraph{Bounding the individual errors.}
Write $M^{\mathrm{gen}}$ for the weighted expected gradient after a generation operation.
Conditional independence of token draws and \Cref{eq:smc-ess-weight-control} give
{\small
\begin{equation}
  \e\!\left[\norm{M^{\mathrm{gen}}-M}^2\mid\mathcal F\right]
    \leq\frac{(1+v)H^2}{N^2}\sum_i\cweight_i^2
    \leq\frac{\rho^2(1+v)H^2}{\tau N}\overline{\cweight}^{\,2}.
\label{eq:smc-local-gradient-variance}
\end{equation}
}
For a triggered systematic resample, applying \Cref{eq:smc-ordered-count-variance,eq:smc-order-regularity} yields
{\small
\begin{equation}
\begin{aligned}
  \e\!\left[\norm{M^{\mathrm{res}}-M}^2\mid\mathcal F\right]
    &=\frac{\overline{\cweight\bweight}^{\,2}}{N^2}
      \e\!\left[\left\|\sum_i(C_i-Np_i)
        \frac{m_\param(s_i)}{\bweight_i}\right\|^2\mid\mathcal F\right]\\
    &\leq\frac{K_{\mathrm{ord}}H^2}{N b_{\min}^2}
        \overline{\cweight\bweight}^{\,2}
     \leq\frac{K_{\mathrm{ord}}\rho^2 H^2}{N}\overline{\cweight}^{\,2}.
\end{aligned}
\label{eq:smc-systematic-gradient-variance}
\end{equation}
}
The final inequality uses $\overline{\cweight\bweight}\leq b_{\max}\overline{\cweight}$.
A skipped check has zero resampling variance and satisfies the same upper bound.
The remaining factor $\overline{\cweight}^{\,2}$ is random and must not be replaced by $1$.
Repeating the same calculation with the constant function $1$ in place of $m_\param$ yields
{\small
\begin{equation}
\begin{aligned}
  \e[(\overline{\cweight}^{\mathrm{gen}})^2\mid\mathcal F]
    &\leq\left(1+\frac{\rho^2 v}{\tau N}\right)\overline{\cweight}^{\,2},\\
  \e[(\overline{\cweight}^{\mathrm{res}})^2\mid\mathcal F]
    &\leq\left(1+\frac{K_{\mathrm{ord}}(\rho-1)^2}{4N}\right)\overline{\cweight}^{\,2}.
\end{aligned}
\label{eq:smc-total-weight-second-moment}
\end{equation}
}
For the second line, apply \Cref{eq:smc-ordered-count-variance} to $z_i=1/\bweight_i$ and use the second inequality in \Cref{eq:smc-order-regularity}.
This bounds the conditional variance added by resampling by $K_{\mathrm{ord}}(\rho-1)^2\overline{\cweight}^{\,2}/(4N)$; the conditional mean remains $\overline{\cweight}$.
In particular, this resampling contribution vanishes when $\bweight=1$, as in the weight-only calculation above.
Let $\overline{\cweight}_k$ denote the mean weight after token position $k$ and its resampling check, if any.
Starting from $\overline{\cweight}_0=1$, applying these bounds successively over the first $k$ token positions gives
{\small
\begin{equation}
  \e[\overline{\cweight}_k^{\,2}]
  \leq\left[
    \left(1+\frac{\rho^2v}{\tau N}\right)
    \left(1+\frac{K_{\mathrm{ord}}(\rho-1)^2}{4N}\right)
  \right]^k
  \leq e^{\frac{k}{N}\left(\rho^2v/\tau+K_{\mathrm{ord}}(\rho-1)^2/4\right)}.
\label{eq:smc-weight-stability-bound}
\end{equation}
}
The same upper bound applies just before resampling at token position $k$.
Substituting these bounds into \Cref{eq:smc-additive-gradient-accounting}, with $l$ generation terms and at most $l-1$ nonzero resampling terms, proves
{\small
\begin{equation}
  \e\!\left[\norm{\hat g_{\mathrm{SMC}}-g}^2\right]
  \leq\frac{\rho^2 H^2 l}{N}\left(\frac{1+v}{\tau}+K_{\mathrm{ord}}\right)
    e^{\frac{l}{N}\left(\rho^2v/\tau+K_{\mathrm{ord}}(\rho-1)^2/4\right)}.
\label{eq:smc-finite-population-variance-bound}
\end{equation}
}
The explicit factor $l$ comes from adding the errors of the $l$ generation steps and at most $l-1$ resampling steps.
ESS controls the generation terms, while the particle-order assumption controls the systematic-resampling terms.

\paragraph{Horizon regime and scope of the bound.}
For the additive-in-length conclusion, assume that $\rho$, $v$, $1/\tau$, and $K_{\mathrm{ord}}$ remain bounded and that, for a constant $c$ independent of $l$ and $N$,
{\small
\begin{equation}
  \frac{l}{N}\left(\frac{\rho^2v}{\tau}
    +\frac{K_{\mathrm{ord}}(\rho-1)^2}{4}\right)\leq c.
\label{eq:smc-horizon-stability}
\end{equation}
}
Then \Cref{eq:smc-finite-population-variance-bound} is $O(H^2l/N)$, rather than the $H^2(1+v)^l/N$ upper bound obtained for independent importance-corrected gradients from $\norm{\reward\nabla_\param\log\bpolicy}\leq H$ and $\e[\cweight^2]\leq(1+v)^l$.
We observe that if $l$ grows very large and $N$ is fixed,  
the SMC bound can still grow exponentially outside \Cref{eq:smc-horizon-stability}.
Because the factor is $O(H^2l/N)$, one also requires that $H$ remains bounded across $l$
for a true linear bound.

For comparison, $\norm{\hat g_{\mathrm{SMC}}}\leq H\overline{\cweight}_l$ and unbiasedness also give the direct bound
{\small
\begin{equation}
  \e\!\left[\norm{\hat g_{\mathrm{SMC}}-g}^2\right]
  \leq H^2\e[\overline{\cweight}_l^{\,2}]-\norm{g}^2.
\label{eq:smc-direct-gradient-bound}
\end{equation}
}
One may take the smaller of this bound and \Cref{eq:smc-finite-population-variance-bound}; the latter exposes the $l/N$ factor when the accumulated error is small.

\section{Uncorrected Enrichment as Success-Filtered SFT}
\label{app:enrichment-sft}

In this section, we characterize RL over 
binary rewards using an
uncorrected enrichment $\qpolicy$
as equivalent up to a scale factor 
to supervised-fine tuning (SFT) on samples from $\qpolicy$ that have reward $1$.

Fix one problem and binary terminal rewards $\reward(\sample)\in\{0,1\}$.
We consider baseline-free REINFORCE, summing token log-probability gradients and averaging over responses, without clipping or KL regularization.
Let $\mu_\param^q=\p_{\sample\sim\qpolicy}(\reward(\sample)=1)>0$ be the enriched policy's success probability.
Filtering its responses to retain only successes gives the distribution
{\small
\begin{equation}
  q_\param^+(\sample)
  =\qpolicy(\sample\mid\reward(\sample)=1)
  =\frac{\qpolicy(\sample)\reward(\sample)}{\mu_\param^q}.
  \label{eq:sft-success-distribution}
\end{equation}
}
The expected uncorrected update therefore satisfies
{\small
\begin{equation}
\begin{aligned}
  u
  &=\e_{\sample\sim\qpolicy}\!\left[
    \reward(\sample)\nabla_\param\log\bpolicy(\sample)
  \right]\\
  &=\mu_\param^q\,
    \e_{\sample\sim q_\param^+}\!\left[
      \nabla_\param\log\bpolicy(\sample)
    \right].
  \label{eq:sft-expected-update}
\end{aligned}
\end{equation}
}
The remaining expectation is the SFT ascent direction on successful enriched responses.
To make this explicit, let $\vartheta$ denote the parameters being optimized while $\param$ denotes the fixed rollout parameters, and define the SFT loss $\mathcal L_{\mathrm{SFT}}(\vartheta;\param)$ by
{\small
\begin{equation}
  \mathcal L_{\mathrm{SFT}}(\vartheta;\param)
  =-\e_{\sample\sim q_\param^+}\!\left[
    \log\pi_\vartheta(\sample)
  \right],
  \qquad
  u=-\mu_\param^q\left.
    \nabla_\vartheta\mathcal L_{\mathrm{SFT}}(\vartheta;\param)
  \right|_{\vartheta=\param}.
  \label{eq:sft-gradient-equivalence}
\end{equation}
}
Thus, uncorrected enriched REINFORCE follows the success-filtered SFT gradient, multiplied by the enriched success probability.
In this sense, uncorrected enrichment trains 
the base policy to imitate the successful responses that enrichment produces, 
just like SFT.

\paragraph{Batch-level equivalence.}
For a batch of $N$ enriched responses, let $K=\sum_{i=1}^{N}\reward(\sample_i)$ be the number of successes.
Whenever $K>0$, the same identity holds for the realized batch:
{\small
\begin{equation}
  \underbrace{\frac1N\sum_{i=1}^{N}
    \reward(\sample_i)\nabla_\param\log\bpolicy(\sample_i)
  }_{\text{uncorrected REINFORCE}}
  =\frac KN\,
  \underbrace{\frac1K\sum_{i:\reward(\sample_i)=1}
    \nabla_\param\log\bpolicy(\sample_i)
  }_{\text{success-filtered SFT ascent}}.
  \label{eq:sft-batch-equivalence}
\end{equation}
}
When $K=0$, REINFORCE produces zero update and there are no successful responses for SFT.
The difference is normalization: REINFORCE averages over all responses, whereas success-filtered SFT averages over successes.
For one problem this changes only the gradient magnitude; if SFT averages successes separately within each problem and then weights problems equally, 
omitting the success-rate factors can also change the aggregate gradient direction 
(similar to how cross-problem gradient aggregation can change the direction of GRPO gradients, 
as in \Cref{app:grpo-distortion}).

\section{Training Setups and Extra Experiments}

We provide detailed descriptions for the controlled experiments in Section 2, 
as well as an additional controlled experiment illustrating the variance reduction of 
SMC from Section 3.

\subsection{Training Setups for Experiments in Section 2}
\label{app:illustrative-setup}

\paragraph{Policy architecture.}
The experiments in \Cref{sec:scale-effect,sec:rotation-effect,sec:variance-effect} use a tabular autoregressive policy: a lookup table followed by a softmax, with no hidden layers.
Its input is the sequence of tokens generated so far (the prefix $s$), and its output is a distribution over the next token $a\in\{0,1,2\}$.
Each prefix has its own row of three trainable logits $\param_{s,a}$, giving $\bpolicy(a\mid s)=e^{\param_{s,a}}/\sum_{b=0}^{2}e^{\param_{s,b}}$; parameters are not shared across prefixes.
Generation starts from the empty prefix, appends one token at a time, and stops after four tokens.
There are therefore $1+3+9+27=40$ possible prefixes at which the model makes a decision, giving $120$ trainable parameters.

\paragraph{Task.}
The policy is trained to generate sequences 
of length $4$
that receive high reward under a fixed scoring rule.
Reward is terminal, given only on the full sample.
A complete sequence $\sample$ receives $\reward(\sample)=1$ 
if it begins with \texttt{000}, $1/2$ if it begins with \texttt{11}, and $0$ otherwise.
For example, \texttt{0002} earns $1$, \texttt{1102} earns $1/2$, and \texttt{0102} earns $0$.
Of the $3^4=81$ possible sequences, three receive reward $1$ 
and nine receive reward $1/2$; 
the variance experiment changes the reward-$1$ prefix to \texttt{00}, as specified below.

\paragraph{Shared training.}
Unless stated otherwise, all logits start at zero, assigning probability $1/3$ to each next token and $1/81$ to each complete sequence.
Updates use the base-policy score $\nabla_\param\log\bpolicy(\sample)$, 
without reward baselines, GRPO normalization, momentum, adaptive optimization, or gradient clipping.
Evaluation uses exact base-policy expected reward, not sampled estimates.

\paragraph{Scale.}
We use the objective $J(\param)=\e_{\sample\sim\bpolicy}[\reward(\sample)]-0.4K(\param)$, where $K(\param)=\kld(\bpolicy\|\pi_{\param_0})$ is computed over complete responses relative to the fixed initial policy.
For each $\alpha\in\{1/16,1/4,1,4,16,64\}$, we multiply the reward gradient by $\alpha$, leave the KL gradient unchanged, and divide the learning rate by $\alpha$.
The resulting update is
{\small
\begin{equation}
  \param_{k+1}=\param_k+0.5\left.
    \nabla_\param\left(
      \e_{\sample\sim\bpolicy}[\reward(\sample)]
      -\frac{0.4}{\alpha}K(\param)
    \right)\right|_{\param=\param_k}
\end{equation}
}
The coefficient 0.5 is the learning rate.
We compute both gradients exactly and run $400$ updates per value of $\alpha$ to 
simulate a training run with a scale-only difference, as stated in \Cref{sec:scale-effect}.

\paragraph{Rotation.}
We set $J(\param)=\e_{\sample\sim\bpolicy}[\reward(\sample)]$
and construct the enriched policy 
$\qpolicy$ by multiplying the base probability of token $1$ by $e^{1.5}$ at the empty prefix and at prefix \texttt{1}, then normalizing the three action weights at each prefix.
All other next-token distributions are unchanged, so the guide favors completing \texttt{11} without assisting the reward-$1$ family.
We compute $g=\e_{\sample\sim\bpolicy}[\reward(\sample)\nabla_\param\log\bpolicy(\sample)]$ and $u=\e_{\sample\sim\qpolicy}[\reward(\sample)\nabla_\param\log\bpolicy(\sample)]$ exactly at each run's current parameters.
Starting from the same uniform policy, the two runs take $700$ updates in their respective 
scale-normalized directions, each with Euclidean step length $0.02$.
Thus neither sampling noise nor differences in update length explain the comparison.

\paragraph{Variance.}
We set $J(\param)=\e_{\sample\sim\bpolicy}[\reward(\sample)]$, and then 
change the reward such that
prefixes with \texttt{00} (instead of \texttt{000}) receive reward $1$.
We then 
initialize the token-$0$ logits at the empty prefix and prefix 
\texttt{0} to $2$, leaving all other logits at zero.
The enriched policy $\qpolicy$ retains the rotation experiment's
$e^{1.5}$ preference toward \texttt{11} and additionally multiplies token-$0$ 
probabilities at the empty prefix and prefix \texttt{0} by $e^{-6}$ before local normalization.
The initial reward-$1$ family probabilities are consequently $0.6193$ under 
$\pi_{\param_0}$ and $3.022\times10^{-5}$ under $q_{\param_0}$.
For each $N\in\{16,32,64,128,256,512,1024\}$, we run $256$ independent trials per $N$, each taking $12{,}670$ updates with learning rate $0.05$ and $N$ fresh responses per update.
The unenriched arm averages $\reward(\sample)\nabla_\param\log\bpolicy(\sample)$ for samples from $\bpolicy$; the corrected arm samples from $\qpolicy$ and multiplies each contribution by $\bpolicy(\sample)/\qpolicy(\sample)$, without clipping or self-normalization, 
to compute the gradient.
An endpoint is counted as suboptimal when its expected reward is below $0.9$, and error bars are pointwise $95\%$ confidence intervals.
Both arms have equal sample budgets at each $N$; increasing $N$ increases the total budget, rather than merely reorganizing a fixed number of samples.

\subsection{Controlled Experiment Illustrating SMC Variance Reduction}
\label{app:smc-experiment}

\paragraph{Setup.}
We use a tabular autoregressive policy with $32$ parameters: $16$ softmax logits for selecting a target string and one match-probability logit per target.
The first token selects one of $16$ fixed target strings; the policy then generates $l$ tokens from a vocabulary of size $16$.
At each position, the policy
emits the corresponding target token with a target-specific probability and distributes the remaining probability uniformly over the other $15$ tokens.
A completed response receives reward $1$ if its number of matching tokens exceeds a threshold and $0$ if it falls below; at the threshold, an independent coin flip determines the binary reward.
We set the threshold and coin probability so that the base success rate is exactly $10^{-5}$ at every length, separating the effect of horizon from changing reward sparsity.
We use $24$ fixed policy/task instances and vary $l\in\{24,32,48,64,80,96,112,128\}$, excluding the initial selection token from $l$.
Unlike the experiments above, we freeze the policy parameters and compare gradient estimates rather than training outcomes.
To construct the enrichment, we compute $1+(e^8-1)\p_{\bpolicy}(\reward=1\mid s)$ at each prefix $s$, then perturb its logarithm with fixed random noise of root-mean-square magnitude $0.5$ to obtain the enrichment weight $\bweight(s)$.
The noise depends on token position and match count, is shared across targets, and leaves the empty prefix and terminal weights unchanged.
The enriched policy $\qpolicy$ multiplies each next token's base probability by the enrichment weight of the resulting prefix and normalizes locally.
Although this is a constructed domain to illustrate the effect of SMC variance reduction, we note
that such random-noise perturbed policies can arise commonly in token-level proposals,
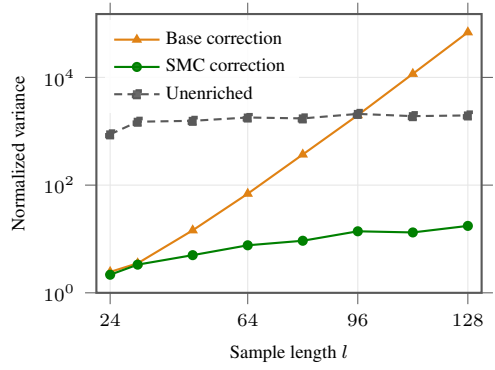
\begin{wrapfigure}{R}{0.48\textwidth}
  \centering
  \begin{tikzpicture}
\begin{axis}[
  compact paper axis,
  width=0.76\linewidth,height=0.55\linewidth,scale only axis,
  /tikz/font=\scriptsize,
  label style={font=\scriptsize},
  xlabel={Sample length $l$},
  ylabel={Normalized variance},
  ymode=log,xmin=20,xmax=132,ymin=1,ymax=150000,
  xtick={24,64,96,128},
  ytick={1,100,10000},
  legend style={at={(0.02,0.98)},anchor=north west,fill=white,fill opacity=0.9,text opacity=1},
]
\addplot[plotuncorrected,mark=triangle*] table[x=T,y=sis,col sep=comma]{figs/data/smc_horizon.csv};
\addlegendentry{Base correction}
\addplot[plotcorrected,mark=*] table[x=T,y=smc,col sep=comma]{figs/data/smc_horizon.csv};
\addlegendentry{SMC correction}
\addplot[black!65,densely dashed,mark=square*] table[x=T,y=iid,col sep=comma]{figs/data/smc_horizon.csv};
\addlegendentry{Unenriched}
\end{axis}
\end{tikzpicture}
  \caption{Normalized variance compared over enriched base correction (as described at the head of
    \Cref{sec:smc}, SMC correction, and unenriched estimators.
  }
  \label{fig:smc-horizon}
\end{wrapfigure}
where constant training makes the parameters of an LLM drift, and subsequently a perfectly 
cancelling enrichment (i.e., a perfect twist as in~\citep{zhao2024probabilistic}) cannot exist.

We compare $N=512$ independent samples from $\bpolicy$, $N=512$ independent samples from $\qpolicy$ with full importance correction, and $N=512$ particles using the dual-weight SMC corrector in \Cref{app:smc-guidance}.
SMC checks the ESS of the product weights $\cweight_i\bweight_i$ after each nonterminal token and resamples when it falls below $0.8N$.
All three estimators subtract the same fixed reward baseline $J(\param)=10^{-5}$; correction weights are neither clipped nor self-normalized, preserving the expected gradient $g$.
We compute $g$ and the two independent estimators' variances exactly by dynamic programming over the selected target, token position, and match count.
For SMC, we estimate variance as mean squared error relative to $g$ over $64$ independent batches per instance and length; MSE equals variance because the estimator is unbiased.

\paragraph{Results.}
\Cref{fig:smc-horizon} shows that independent correction and SMC have similar variance at short horizons, but diverge sharply as the responses grow longer.
The variance of independent correction is $1.13$ times that of SMC at $l=24$, rising to approximately $3{,}929$ times at $l=128$.
At the longest horizon, independent correction has approximately $35$ times the variance of unenriched sampling, whereas SMC's variance is approximately $1/112$ that of unenriched sampling.


\end{document}